\documentclass[sn-mathphys]{sn-jnl}

\jyear{2021}%

\theoremstyle{thmstyleone}%
\theoremstyle{thmstyletwo}%

\theoremstyle{thmstylethree}%

\newcommand{\Acondition}{\emph{CBCL} condition}
\newcommand{\Bcondition}{\emph{FT} condition}
\newcommand{\Ccondition}{\emph{JT} condition}

\usepackage{caption}
\usepackage{subcaption}
\usepackage{comment}
\usepackage{amsmath}
\usepackage{enumerate}
\usepackage[shortlabels]{enumitem}
\usepackage{array}
\newcolumntype{P}[1]{>{\centering\arraybackslash}p{#1}}

\begin{document}

\title[Human Perceptions of a Continual Learning Robot]{
Continual Learning through Human-Robot Interaction: Human Perceptions of a Continual Learning Robot in Repeated Interactions}


\author*[1,2]{\fnm{Ali} \sur{Ayub}}\email{ali.ayub@concordia.ca}

\author[3]{\fnm{Zachary} \sur{De~Francesco}}\email{zdefrancesco@uwaterloo.ca}

\author[4]{\fnm{Patrick} \sur{Holthaus}}\email{p.holthaus@herts.ac.uk}

\author[2,3,4]{\fnm{Chrystopher} \sur{L.~Nehaniv}}\email{chrystopher.nehaniv@uwaterloo.ca}

\author[3,4]{\fnm{Kerstin} \sur{Dautenhahn}}\email{kerstin.dautenhahn@uwaterloo.ca}

\affil*[1]{\orgdiv{Concordia Institute for Information Systems Engineering (CIISE)}, \orgname{Concordia University}, \orgaddress{
\city{Montreal}, \postcode{H3G 1M8}, \state{Quebec}, \country{Canada}}}

\affil*[2]{\orgdiv{Department of Systems Design Engineering}, \orgname{University of Waterloo}, \orgaddress{
\city{Waterloo}, \postcode{N2L 3G1}, \state{Ontario}, \country{Canada}}}

\affil[3]{\orgdiv{Department of Electrical and Computer Engineering}, \orgname{University of Waterloo}, \orgaddress{
\city{Waterloo}, \postcode{N2L 3G1}, \state{Ontario}, \country{Canada}}}

\affil[4]{\orgdiv{School of Physics, Engineering and Computer Science}, \orgname{University of Hertfordshire}, \orgaddress{
\city{Hertfordshire}, \postcode{AL10 9AB}, \state{England}, \country{UK}}}


\abstract{For long-term deployment in dynamic real-world environments, assistive robots must continue to learn and adapt to their environments. Researchers have developed various computational models for continual learning (CL) that can allow robots to continually learn from limited training data, and avoid forgetting previous knowledge. While these CL models can mitigate forgetting on static, systematically collected datasets, it is unclear how human users might perceive a robot that continually learns over multiple interactions with them. In this paper, we developed a system that integrates CL models for object recognition with a Fetch mobile manipulator robot and allows human participants to directly teach and test the robot over multiple sessions. We conducted an in-person study with 60 participants that interacted with our system in 300 sessions (5 sessions per participant). We conducted a between-subject study with three different CL models to understand human perceptions of continual learning robots over multiple sessions. Our results suggest that participants' perceptions of trust, competence, and usability of a continual learning robot significantly decrease over multiple sessions if the robot forgets previously learned objects. However, the perceived task load on participants for teaching and testing the robot remains the same over multiple sessions even if the robot forgets previously learned objects. Our results also indicate that state-of-the-art CL models might perform unreliably when applied on robots interacting with human participants. Further, continual learning robots are not perceived as very trustworthy or competent by human participants, regardless of the underlying continual learning model or the session number.}



\maketitle
\section{Introduction}
\label{sec:introduction}
Assistive robots are becoming an integral part of our society in a variety of roles, such as caregivers, cleaning robots, or home assistants \cite{Matari17,petrecca_how_2018,Saunders16,Koay21,reiser13,shah_hri_23}. However, for robots to be able to assist people in daily environments over a long period of time, they must adapt to the changing needs of their users and their environments. As it would be impossible to pre-program all the tasks a robot needs to perform and all the items a robot might encounter in a person's daily environment, robots will need to continually learn interactively on the fly from their users who are likely unfamiliar with robotics and machine learning (ML).  

To operate in daily environments, a general task for a robot is to learn and understand the objects in its environment \cite{Dehghan19,Valipour17,Ayub_IROS_20,thomaz09}. Such a task is central to a variety of different service tasks such as fetching and carrying objects, cooking and meal preparation, doing the dishes and the laundry, etc. Various machine learning models have been developed in the last decade for achieving remarkable performance on object recognition tasks \cite{He_2016_CVPR,Simonyan14}. However, one of the main challenges faced by robots using ML models to continually learn objects is \textit{catastrophic forgetting} \cite{french19,mcclelland95}. Catastrophic forgetting\footnote{Note that the term catastrophic forgetting is mainly used in Machine Learning literature to describe the phenomenon of an ML model forgetting most past knowledge on static datasets. However, when interacting with real users, the perception of forgetting might be far from ``catastrophic''.} occurs when a continual learning (CL) agent forgets the previously learned knowledge when learning new information \cite{Rebuffi_2017_CVPR}. This, however, is in contrast with human memory that might gracefully forget detailed experiences but keeps abstract knowledge consolidated in long-term memory \cite{costanzi21,Mack18}. One approach to avoid forgetting is to store the data of the previously learned tasks in memory and retrain the CL agent on the previously stored data plus the new information. However, this can lead to computational processing and memory storage issues for real-world robots with real-time constraints, limited onboard memory, and computational resources. In recent years, different research directions (some inspired by neuroscience \cite{kirkpatrick17,kemker18}) have been taken in the field of continual machine learning to mitigate the catastrophic forgetting problem without storing and relearning the complete dataset of the previous tasks~\cite{mudt2020,mundt2022clevacompass,Smith_2021_ICCV,lomonaco17,Chaudhry_2018_ECCV,hayes20,smith2022incremental}. While state-of-the-art (SOTA) CL models alleviate catastrophic forgetting, they still suffer from some forgetting when learning over a large number of repeated sessions \cite{Rebuffi_2017_CVPR,Li18,kirkpatrick17,ayub2021eec,Ostapenko_2019_CVPR}.

Another challenge faced by continual learning robots is that their users might not provide a sufficiently large number of data (examples) to train an ML model. In the past few years, robotics researchers developed CL models that can learn continually from only a few training examples per object, while also mitigating catastrophic forgetting \cite{Ayub_2020_CVPR_Workshops,Tao_2020_CVPR,lesort2020continual}. This problem is known as Few-Shot Incremental Learning (FSIL) \cite{Ayub_2020_CVPR_Workshops,Tao_2020_CVPR,tao_2020_ECCV,Zhang_2021_CVPR}. Although FSIL approaches have produced promising results on systematically collected ``non-social'' datasets by the experimenters, it is unknown how these systems might perform when learning from human participants. Further, it is also unknown how people might perceive robots that continually learn through interaction with their users. 

For the long-term deployment of robots in human environments, it is critical that we understand how humans might perceive such robots, as these robots will interact with and operate around humans~\cite{Koay21,paetzel20,lyons2023explanations,de2020towards}. Trust is one of the essential components for people's relationships with autonomous robots~\cite{rossi17,nayyar18,andras2018trusting,esterwood2021you}. Most prior research showed that people lose trust in autonomous robots that might be imperfect. Research also showed that people's trust in robots could strengthen over time if they are involved in teaching the robot~\cite{chi_hri_23}, however, this research was conducted with a simulated robot that did not actually learn from user instructions and used pre-programmed behavior. Other important factors that influence people's relationships with robots over the long term are perceptions of social attributes and usability of robots~\cite{scheunemann22,Koay21,scassellati2018improving,de2016long,de2017they}. Finally, for autonomous robots that learn through human teaching, it is also imperative to understand human perceptions of task load for interacting with and teaching the robot~\cite{kosch_2023,zhang2022trans4trans}. Most of the prior research on analyzing human perceptions of robots has been conducted in a single interaction scenario, using hand-crafted, heuristic approaches. To the best of our knowledge, we know of no other work on testing CL or FSIL models deployed on robots that directly learn from human users on the fly over multiple interactions, which is the focus of this paper.

Here, we consider a system for socially guided continual learning (SGCL) and conduct an in-person user study to explore how people perceive a robot that continually learns common household objects over multiple interactions. We developed a system that integrates a graphical user interface (GUI) on an Android tablet with a CL model deployed on the Fetch mobile manipulator robot \cite{Wise16}. In this system, we focused solely on the continual learning of objects and avoided adding any extra social behaviours to the robot that might affect human perceptions of the robot. We performed a long-term between-participant user study (N=60) where participants interacted with a fully autonomous Fetch mobile manipulator robot that used three different CL models: one that suffers from forgetting on static ``non-social'' datasets, another state-of-the-art (SOTA) approach for FSIL that mitigates forgetting, and the upper bound approach that stores and retrains all of the previous data when learning new information. We conducted 300 interactive sessions with 60 participants, where each participant taught 25 household objects to the robot in 5 sessions with 5 objects per session. We used four questionnaires in the study to answer the following research questions:

\begin{itemize}[leftmargin=0.4in]
    \item [\textbf{RQ1}] How do human perceptions of trust, social attributes, task load, and usability evolve when interacting with a continual learning robot over multiple sessions?
    \item [\textbf{RQ2}] Is there a difference in participants' perceptions of trust, social attributes, task load, and usability of a continual learning robot for different continual learning models? 
\end{itemize}

The remainder of the paper is organized as follows: Section \ref{sec:related_work} reviews related work including background on continual learning and prior continual learning approaches, prior work on robot learning from human teachers, and research on perceptions of robots. Section~\ref{sec:methodology} explains our unique socially guided continual learning setup and the continual learning models used in the study. Section~\ref{sec:method} describes the hypotheses guided by the RQs, participants recruited for the study, experimental setup and the procedure, and the measures used to evaluate the data collected from the study. Section~\ref{sec:results} describes the detailed results of the study, followed by Section~\ref{sec:discussion} to discuss the implications of the results. Finally, Section~\ref{sec:conclusion} concludes the article, followed by Section~\ref{sec:limtiations} that discusses the limitations of the study and directions for future research.

\section{Related Work}
\label{sec:related_work}
In this section, we first present an overview of continual learning and ML methods for continual learning that are mostly tested without human users, followed by methods for robot learning that are designed to learn from human users, albeit in a single interaction. We then describe research on evaluating human perceptions of robots over single and multiple interactions.

\textit{Continual Learning.} The standard continual learning (CL) problem for an object recognition task is defined as: Suppose a CL model $\mathcal{M}$ gets a stream of labeled training datasets $D^1, D^2, ...$ over multiple increments, where $D^t=\{x_i^t,y_i^t\}_{i=1}^{\lvert D^t \rvert}$ is the dataset in the $t$th increment, $x_i^t$ is the $i$the data point in $D^t$ with label $y_i^t$. $L^t$ is the set of object classes in the $t$th training dataset, where $L^j \cap L^k =\varnothing, \forall j\neq k$. During the testing phase, if the model $\mathcal{M}$ is given the increment label when predicting the class label of a data point, this setup is known as task-incremental learning \cite{Chaudhry_2018_ECCV,Li18,kirkpatrick17}. In contrast, for the class-incremental learning (CIL) setup, the model $\mathcal{M}$ is tested in increment $t$ on data points belonging to any of the previous classes ($L^1,...,L^t$) without access to the increment label \cite{Wu_2019_CVPR,Castro_2018_ECCV,Rebuffi_2017_CVPR,Kang_2022_CVPR}. CIL is a more realistic continual learning setup, as robot users might not be willing to (or even remember) the increment label when asking the robot to predict the class label of an object. Therefore, we mainly review CIL approaches in this paper.    

\textit{Class-Incremental Learning.} Various research directions have been taken in the past to develop CIL models that can mitigate the catastrophic forgetting problem \cite{Rebuffi_2017_CVPR,Castro_2018_ECCV,Hou_2019_CVPR,Wu_2019_CVPR,Ayub_2020_CVPR_Workshops}. Most existing class-incremental learning (CIL) methods avoid catastrophic forgetting by storing a portion of the training samples from previous classes and retraining the model on a mixture of the stored data and new data \cite{Rebuffi_2017_CVPR,Castro_2018_ECCV,Wu_2019_CVPR,hayes2021replay}. However, this approach does not scale as additional data exhausts memory capacity limiting performance in real-world applications. To avoid this problem, some CL approaches use regularization techniques \cite{Li18,kirkpatrick17}. Although these approaches solve the memory storage issues, their performance is significantly inferior to approaches that store old class data. Another set of approaches uses generative replay to avoid storing raw data and generate old data using stored class statistics \cite{ayub2021eec,Ostapenko_2019_CVPR,kemker18,Shin17,Wu18_NIPS}. Generative approaches, however, do not scale well to learning over longer sequences and their performance deteriorates drastically. One of the major concerns for all CIL approaches is that they perform poorly when learning from limited training data \cite{Ayub_2020_CVPR_Workshops,Tao_2020_CVPR}. Therefore, they are not suitable for learning from human users who might be unwilling to provide hundreds or thousands of images per object class.

\textit{Few-Shot Incremental Learning.} In the past couple of years, CL researchers developed class-incremental learning models that continually learn from a small number of training examples per class. This setup is known as few-shot incremental learning (FSIL) \cite{Ayub_2020_CVPR_Workshops,Tao_2020_CVPR}. All of these approaches train a CL model on a large number of object classes (called base classes) with a large dataset in the first increment to learn a good representation of the data. In the next increments, the model utilizes the representation learned in the first increment to learn new classes with only a few training images per class \cite{Zhang_2021_CVPR,Tao_2020_CVPR,tao_2020_ECCV,Ayub_2020_CVPR_Workshops,Bhunia_2022_CVPR,Hersche_2022_CVPR}. These approaches, however, were only tested on static, simple datasets (e.g. MNIST \cite{Lechun98}) and not on real robots that might not have perfect data available. Although a few CL approaches \cite{Ayub_IROS_20,Dehghan19,Valipour17} have been tested with robots in recent years, most of them have been tested on only a small number of object classes (usually 10 or fewer). Further, none of these FSIL approaches were tested with real participants, and the data was captured by the experimenters in systematically controlled setups. Non-expert users, however, might not be aware of the underlying CL models, therefore they might provide imperfect data to the robot during teaching (or testing). It is also unclear how human participants might perceive CL systems on robots, and if they consider such systems to be feasible and easy to use.

\textit{Robot Learning with Human Teachers.} A few studies have been conducted in the past with human participants to teach robots different manipulation tasks \cite{bobu21} or object classes \cite{thomaz09,chao10}. For example, Bobu et al. \cite{bobu21} developed a reinforcement learning technique for a manipulator robot that can perform simple manipulation tasks with human assistance. Thomaz et al. \cite{thomaz09} developed an object learning system that allowed a robot to learn object names and simple affordances from interactions with human participants. Human participants taught 6 simple objects to a social robot, which used a support vector machine (SVM) based method for learning these objects. Thomaz et al. showed that there were significant performance differences when machine learning models learned from human teachers rather than using systematically collected object datasets. Although these studies developed and tested ML techniques with human teachers, they were only tested in a single interaction with the users. Note that a single teaching interaction with the robot might not be a correct indicator of human perceptions of a continual learning robot, as user perceptions might change when teaching and testing the robot for the same task over multiple interactions (see Section \ref{sec:results} for results). In addition, single-session research with robots can suffer from the novelty effect which is well known in HRI research~\cite{leite2009time,babel2022will,sung2009robots}.

\textit{Perceptions of Robots.} For the long-term deployment of robots in human environments, it is critical that we understand how humans might perceive such robots \cite{Koay21,paetzel20,lyons2023explanations,de2020towards}. Humans ascribe social traits and meaning to any agent in motion \cite{Hoffman14}, but controlled experiments can help understand humans' perceptions about different aspects of robots \cite{anzalone__2015,takayama11,aliasghari21}. For example, different studies have been conducted in the past that evaluate humans' trust in social robots exhibiting erroneous behaviors \cite{rossi17,nayyar18,robinette17,rossi20}. Most of these studies indicated that the robot's performance on different tasks might affect humans' trust in the robot. Other studies analyzed human perceptions of social robots in terms of competence and warmth for a variety of tasks \cite{scheunemann22,paetzel20}. Studies have also been conducted for domestic and industrial robots to understand their perceived usability \cite{Koay21,gadre19,solanes20,louie20}. Although a few of these studies have been conducted to understand long-term interaction (maximum three sessions) with robots \cite{rossi20,paetzel20}, these studies used hand-crafted, heuristic approaches, and not any modern CL approaches. 
Unlike these prior research, a unique aspect of our study is that the robot directly learns new objects through human teaching and then uses the learned knowledge to autonomously find objects.

\begin{figure}[t]
\centering
\includegraphics[width=\linewidth]{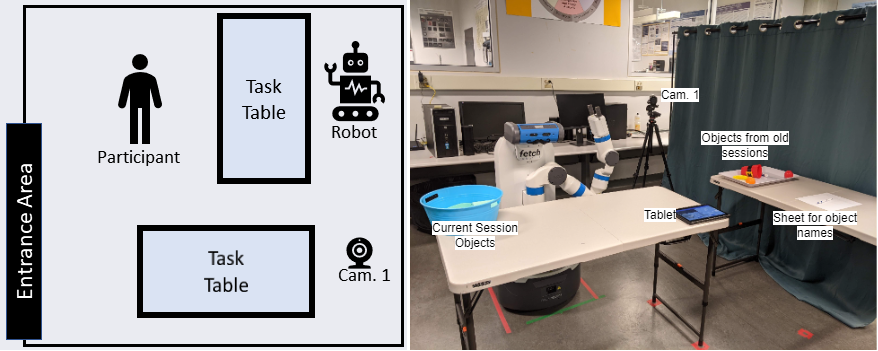}
\caption{\small (Left) Experimental layout for the SGCL setup with the participant and the robot. (Right) Corresponding  real-world setup.}
\label{fig:experimental_setup}
\end{figure}

Recently a few studies analyzed human perceptions of robots learning directly from human demonstrations. For example, Schrum et al.~\cite{schrumtowards} tested how human perceptions changed when the robot provided feedback related to human teaching. This approach, however, was only tested in a single session and was not used to test a continual learning approach. Liu et al. \cite{Liu_2020} presented a framework for learning online from human users but did not test their approach with real participants. 

In contrast to previous work, to the best of our knowledge, we conducted the first user study at the intersection of continual learning and HRI, to understand human perceptions of a robot regarding various aspects (trust, competence, system usability), when the robot continually learns from human users over five sessions. 

\section{Socially Guided Continual Learning}
\label{sec:methodology}
We studied human perceptions of a continual learning robot in the context of an object recognition task. In this setup, the robot learns household objects from the user (in multiple sessions) on a table-top environment, and then finds and points to the requested object on the table after learning them from the user. Figure \ref{fig:experimental_setup} shows the table-top experimental setup for this study. The simplicity of the setup and the task makes it clear what the user should do to teach the robot different objects and what the robot should do to find the learned objects during the testing phase.  

\begin{figure}[t]
\centering
\includegraphics[width=\linewidth]{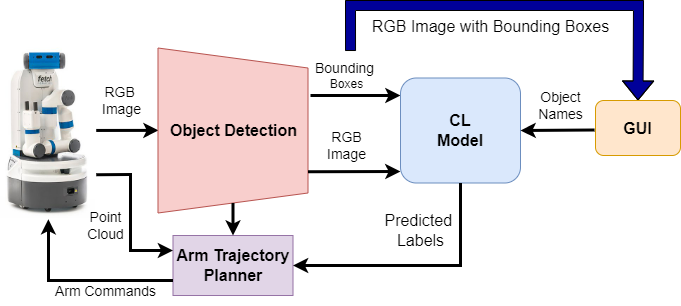}\caption{\small Our complete SGCL system. Processed RGB images from robot's camera are sent to the GUI for transparency and also passed on to the CL Model. The user sends object names to the CL model either for training the CL model or finding an object. The arm trajectory planner takes point cloud data, processed RGB data, and predicted object labels from the CL model as input and sends the arm trajectory for the Fetch robot to point to the object.}
\label{fig:sgcl_architecture}
\end{figure}

For this setup, we consider a socially guided continual learning (SGCL) system for the object recognition task, which integrates continual learning (CL) models with the robot for interactive and transparent learning from human users. Figure \ref{fig:sgcl_architecture} shows the SGCL system for the object recognition task. In this system, in each session (or increment) $t$ the user interacts with the robot through a graphical user interface (GUI) to teach the robot $L_t$ number of objects. The robot captures images of the $L_t$ objects and pre-processes them, getting the labels of the processed object images from the user to generate a dataset $D^t = \{x_i^t,y_i^t\}_{i=1}^{\lvert D^t \rvert }$, where $x_i^t$ is the $i$th image in the dataset with the class label $y_i^t$. The CL model $\mathcal{M}$ then trains on the dataset $D^t$. Note that unlike static CL setups (such as FSIL~\cite{Ayub_2020_CVPR_Workshops}), the number of objects per object class in a session is not fixed as it is dependent on the number of times the user teaches an object to the robot. Further, there can be an overlap in the object classes taught in different sessions depending on how the user labels the objects i.e. $L^j \cap L^k \neq \varnothing$, for any $j\neq k$. For example, the user can name two different cups in different sessions with different names, such as ``green cup'' and ``red cup'', or they can name both of the cups as ``cup''. Such labelling differences among users are generally considered a problem (labelling bias) when developing general-purpose ML models trained offline on large datasets. However, our study focuses on personalized autonomous robots that will learn continually from their individual users about objects in their unique environments. Therefore, we did not constrain the users to teach the robot in the exact same way. 

In the testing phase, the robot receives the request from the user through the GUI to find an object. The robot passes the pre-processed images to the CL model to get the predicted object labels. If the object is found, the robot finds the 3D location of the object on the table and points to the object using its arm. Note that the user has flexibility in terms of the total number of objects to be tested in an increment, as well as which objects to test (old or new objects). Therefore, unlike static CL setups, the test set of objects is not fixed in each session. Due to this flexibility in SGCL, results for CL models were quite different from the results on static datasets (see section \ref{sec:results} for details). It was important for us to introduce this flexibility for the user, to make the setup more similar to real-world scenarios.

\subsection{Continual Learning Models}
The main goal of our study is to do an in-depth analysis of how users perceive CL models over repeated, long-term interactions. To do such an analysis, it is important to choose a meaningful baseline. The naive finetuning (FT) approach \cite{Rebuffi_2017_CVPR} has been used extensively in CL literature as a baseline on static datasets. Therefore, we chose to test FT as our study’s baseline model. The FT approach uses a convolutional neural network (CNN) \cite{He_2016_CVPR} that is trained on the image data of the object classes in each increment (i.e.\ in an interactive session with the user). The model does not train on any of the objects learned in the previous increments (sessions) and therefore it forgets the previously learned objects. More details on this model can be found in~\cite{Rebuffi_2017_CVPR,Ayub_IROS_20}. Please see Section~\ref{sec:ft_justify} for further details on the choice of FT as a baseline for our study.

For the second model, we consider a SOTA CL approach specifically designed for FSIL in robotics applications \cite{Ayub_2020_CVPR_Workshops}. This approach, termed centroid-based concept learning (CBCL), uses a CNN pre-trained on the ImageNet dataset \cite{Russakovsky15} as a feature extractor for object images. In each increment, $t$, the model receives a small number of images for some object classes and extracts feature vectors for the object images using the pre-trained feature extractor. CBCL then clusters the feature vectors of all the object classes in the increment and generates a set of centroids $C^y = \{c_1^y, ..., c_{n_y}^y\}$ for each object class separately, where $n_y$ is the total number of centroids for class $y$. CBCL avoids forgetting by generating separate centroids for each class in a new increment $t$, without changing the centroids of the previously learned classes. 
For the classification of a new object, CBCL finds the distance of the feature vector of the test object from the centroids of all the object classes. CBCL then uses a weighted voting scheme to find the most common class among the closest centroids to the test feature vector. The most common class is predicted as the object class for the test feature vector. More details about CBCL can be found in \cite{Ayub_2020_CVPR_Workshops}. CBCL has been shown to produce promising results when learning from systematically collected object datasets by experts (researchers). However, it was never trained or tested in real-time with human participants. 

Finally, for the third model, we consider the batch learning (called joint training (JT) in this paper) approach that stores the image data of the object classes from previous increments and retrains using stored data when learning new objects. JT has been used in CL literature as an upper bound for continual learning on static datasets. JT trains a CNN model on a combined image dataset of the new and old object classes in each increment, and therefore its training time continues to increase with each increment. More details about this model can be found in~\cite{Rebuffi_2017_CVPR,Ayub_IROS_20}. 
For our study, similar to \cite{Ayub_IROS_20}, we used a CNN pre-trained on the ImageNet dataset instead of using a CNN with random weights for FT and JT, to enable learning from a few training examples. 
In this paper, we integrate FT, CBCL, and JT in a fully autonomous system that allows users to experience these different ML models in real time through the Fetch mobile manipulator robot \cite{Wise16}. 

\section{Method}
\label{sec:method}
To answer the two research questions (RQ1,2 in Section \ref{sec:introduction}), we tested different hypotheses, related to those research questions, in a repeated measures study where users interacted over five sessions with the system (Section \ref{sec:methodology}). 

\subsection{Hypotheses}
\label{sec:hypotheses}
The following hypotheses are guided by previous research that was  discussed in Section \ref{sec:related_work}:  Prior HRI research showed that users' perceptions of trust, usability, and social attributes are correlated with the performance of the robot, whereas the perceptions of task load are correlated to the time and effort spent in interacting with the robot. Further, prior CL research showed that CL models can forget previous knowledge over time, and thus their performance decreases. However, there is a difference in the rate of forgetting for different CL models. 

Note, H$n.m$ is the $m$th hypothesis related to the research question $n$, e.g. H1.3 is the third hypothesis to answer RQ1.

\noindent \textit{Trust}
\begin{itemize}[leftmargin=0.5in]
        \item [\textbf{H1.1}]
        Users' perceptions of trust decrease in the robot over multiple sessions regardless of the CL model.
        \item [\textbf{H2.1}] A robot that forgets is perceived as less trustworthy than a robot that remembers most previously learned objects.
        \item [\textbf{H2.2}] A robot that retrains on all previous objects is perceived as more trustworthy than a robot that does not retrain on all previous objects. 
\end{itemize}
\noindent \textit{Social Attributes}
\begin{itemize}[leftmargin=0.5in]
        \item [\textbf{H1.2}]
        Users' perception of the social attributes of the robot decreases over multiple sessions regardless of the CL model.
        \item [\textbf{H2.3}] The social attributes of a robot that forgets are perceived to be worse than those of a robot that remembers most previous objects.
        \item [\textbf{H2.4}] The social attributes of a robot that retrains on all previous objects are perceived to be better than those of a robot that does not retrain on all previous objects. 
\end{itemize}
\noindent \textit{Task Load}
\begin{itemize}[leftmargin=0.5in]
        \item [\textbf{H1.3}]
        Users' perception of the task load for teaching the robot increases over multiple sessions regardless of the CL model.
        \item [\textbf{H2.5}] The task load for teaching and testing a robot that forgets is less than a robot that remembers most previous objects.
        \item [\textbf{H2.6}] The task load for teaching and testing a robot that retrains on all previous objects is more than a robot that does not retrain on all previous objects. 
\end{itemize}
\noindent \textit{Usability}
\begin{itemize}[leftmargin=0.5in]
        \item [\textbf{H1.4}]
        Users' perceptions of the usability of the robot decrease over multiple sessions, regardless of the CL model.
        \item [\textbf{H2.7}] Users perceive a robot that remembers most previous objects to be more useful and easier to use than a robot that forgets.
        \item [\textbf{H2.8}] Users perceive a robot that retrains on all previous objects to be more useful and easier to use than a robot that does not retrain on all previous objects. 
\end{itemize}

\noindent
\subsection{Fetch Mobile Manipulator Robot}
\label{sec:fetch}
Manipulator robots with an RGB-D camera are well-suited for recognizing and manipulating objects. In our setup (Figure \ref{fig:experimental_setup}), we use the Fetch mobile manipulator robot \cite{Wise16}. Fetch consists of a mobile base and a 7 DOF arm. The robot also contains an RGB camera, a depth sensor, and a Lidar sensor. These sensors can be used for 3D perception, SLAM mapping, and obstacle detection in the robot's environment. In our setup, we do not ask the robot to manipulate objects or move its base, allowing us to solely focus on continual learning which is principally about learning and recognizing objects. We mainly use the RGB-D camera to recognize objects and the 7 DOF arm to point to objects. We use ROS packages available with the Fetch robot for moving the torso, and the arm of the robot. We did a safety analysis of the robot (approved by our University's ethics review board), and also adopted several mitigating strategies. Therefore, the robot was considered safe to be used with human participants in our study.

\begin{figure}[t]
\centering
\includegraphics[width=00.8\linewidth]{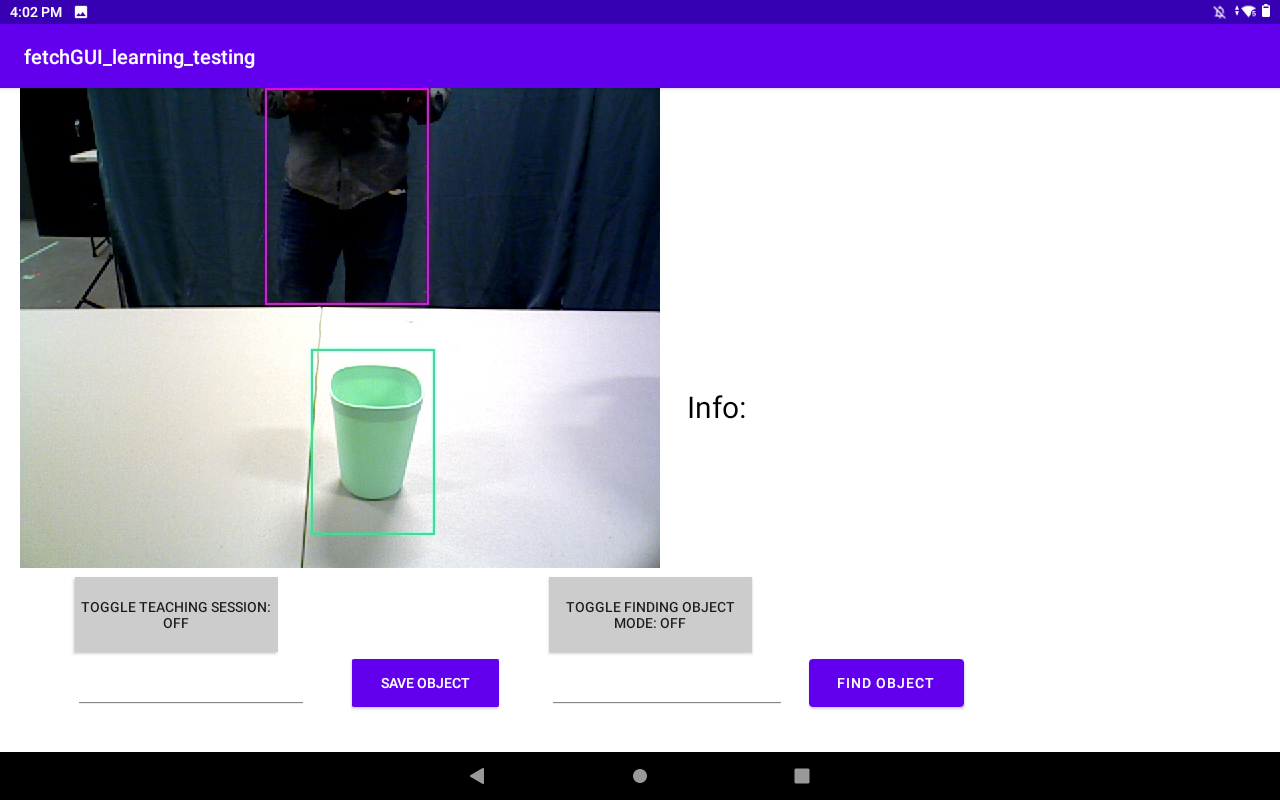}
\caption{\small The graphical user interface (GUI) used to interact with the robot. The RGB camera output with bounding boxes is on the top left. The buttons at the bottom can be used to teach objects to the robot and ask it to find objects in the testing phase. The top right of the GUI shows information sent by the robot to the user.}
\label{fig:gui}
\end{figure}

As there can be multiple objects on the table in front of the robot's camera, we process the RGB images further by passing them through a generic object detector \cite{Redmon_2016_CVPR}. The object detector finds regions in the image that are likely to contain objects (Figure \ref{fig:gui}). The detected regions are filtered using non-max suppression \cite{Hosang_2017_CVPR} to remove any overlaps. We also filter out the detected objects that are not on the table (background objects, participant interacting with the robot as seen in Figure \ref{fig:gui}) using the depth perception of the objects. The resulting regions are cropped into separate images for objects detected on the table and then forwarded to the CL model. 

\subsection{Graphical User Interface}
\label{sec:gui}
For users to be able to interact and teach the robot different objects in an open-ended manner, we created a simple graphical user interface and deployed it on an Android tablet. Figure \ref{fig:gui} shows a screenshot of the GUI. The top left side of the GUI shows the pre-processed camera output of the robot which contains bounding boxes for detected objects. The camera output was used as a transparency device so that the participants could clearly understand what the robot was seeing on the table. On the bottom left of the GUI, there is a toggle button that can be used to start a teaching session with the robot. Once the button is pressed, it turns green indicating that the system was in the teaching phase. After starting the teaching phase, participants can type the name (class label) of the objects in the space below the toggle button. Participants can save an image of the object using the save button next to the empty space. The bottom right of the GUI contains another toggle button that can be used by the participants to start the testing phase. The button turns green once pressed. During the testing phase, participants can type the name (class label) of the object to be found on the table in the space below the testing toggle button. Participants can then press the Find Object button next to the empty space to ask the robot to find the requested object on the table. Finally, the top right section of the GUI shows the messages communicated by the robot to the user during the session. The robot also spoke these messages using a text-to-speech module available in ROS. 

Note that we did not use an NLP (natural language processing) based interaction system because designing an open-ended NLP system for teaching objects is a challenging problem \cite{chai18_ijcai}, and it is prone to additional errors during the speech-to-text and natural language understanding (NLU) phases. These errors could distract participants from the main research goal of the study (continual learning of household objects). Our goal was to study participants' perceptions of the continual learning system, and not to evaluate the communicative capabilities of the robot. 

\subsection{Participants}
\label{sec:participants}
We recruited 63 participants (35 female (F); 28 male (M), all students) from the University of Waterloo, between the ages of 18 and 37 years ($M=23.12$, $SD=4.04$). Out of the 63 participants, three dropped out before completing the study. Out of the 60 participants, all three conditions were randomly assigned 20 participants each (ages: $M=22.7$, $SD=4.58$, 9 F, 11 M for \Acondition{}, ages: $M=24.53$, $SD=4.03$, 10 F, 10 M for \Bcondition{}, ages: $M=22.2$, $SD=2.96$, 15 F, 5 M for \Ccondition{}). Based on their self-assessments in a pre-experiment survey, 35\% of the participants reported that they were familiar with robot programming, 55\% reported that they had previously interacted with a robot, 5\% were familiar with the Fetch robot, and 8\% had previously participated in an HRI study. For the rest of the paper, we call participants with prior robot programming experience `experts' and the rest of the participants `non-experts'. All procedures were approved by the University of Waterloo Human Research Ethics Board.

\subsection{Procedure}
\label{sec:procedure}
We conducted five repeat sessions (each $\sim$20-30 minutes) with each participant in a robotics laboratory. All sessions were video recorded. Each participant was randomly assigned to one of the three experimental conditions using one of the three CL models (CBCL, finetuning, and joint training). Before their first session, each participant was asked to complete a consent form and a pre-experiment survey online (using Qualtrics \cite{qualtrics}). After completing the consent form and the pre-experiment survey, the experimenter greeted the participant and gave a brief oral introduction to the experiment. Participants were told to consider Fetch as their personal household robot and they can teach the robot household objects over five days such that the robot learns 25 objects over time. Next, the participant was directed to the study area of the lab and was handed an Android tablet with the GUI (see Figure \ref{fig:gui}) loaded. The participant was told that they would first run a demo session with the robot to understand how to teach and test the robot. The experimenter also mentioned that during the demo phase, the robot would not be learning any of the objects shown in front of the camera. 

The experimenter explained that each session with the robot will consist of a teaching phase and a testing (i.e. finding an object) phase. The experimenter then used a blue cup as a demo object (this object was not used later) and placed it on the table. The experimenter then asked the participant to stand in the designated area in front of the table and start a teaching session by pressing the ``Toggle Teaching Session'' button on the GUI. Once the button was pressed, it turned green, and the robot sent a message on the tablet stating, ``Entered teaching mode. You can now start teaching me objects.'' The robot spoke the same message through its speakers. The experimenter then asked the participant to type the name of the object in the text box below the toggle button. The participant was also told that they can name the object whatever they like. After the participant named the object, they were told that they can save the object by pressing the ``Save Object'' button next to the text box. Once the ``Save Object'' button was pressed, the robot stated, ``[OBJECT NAME TYPED IN THE TEXT BOX] has been saved''. The experimenter then mentioned to the participant that they can save each object as many times as they want by placing the object at different places on the table at various angles. They further mentioned that a similar procedure can be used to teach the other four objects in the session. The participant was then told that once they are finished saving all five objects in a session, they can press the toggle button again to end the teaching session. 
Once the button was pressed, it turned grey, and the robot stated, ``I am learning the objects, please wait.'' The robot then stated, ``Left teaching mode''. The experimenter explained to the participant that the robot would learn the objects shown by them in the session and then communicate to them when it had finished learning and left the teaching mode. 

The experimenter then explained the testing phase to the participant. The experimenter asked the participant to press the ``Toggle Finding Object Mode'' button to start the testing phase of the session. Once the button was pressed, it turned green and the robot stated, ``Entered finding mode''. The experimenter then placed two other objects on the table alongside the demo object (a total of three objects on the table). The experimenter then mentioned that during the testing phase, the participant can place one or up to three (a suggestion, not a requirement) objects on the table. The experimenter further mentioned that after placing the objects, the participant can type the name of the object to be found by the robot in the text box below the toggle button. The experimenter then asked the participant to type ``cup'' in the text box to ask the robot to find this object on the table. Once they finished typing, the experimenter asked the participant to press the ``Find Object'' button. After the participant pressed this button, the robot stated, ``I will point to the cup now. Please make sure that you are at a safe distance from me.'' The robot then moved its torso and arm to point to the cup on the table, and after finishing it stated, ``I am done''. The experimenter then mentioned that the participant could ask the robot to find the objects taught in the current session and in the previous sessions by placing them on the table and using the ``Finding Object Mode''. The experimenter then asked the participant to press the ``Toggle Finding Object Mode'' button again to leave the testing phase. Once the button was pressed, it turned grey and the robot stated, ``Left finding mode''.

After the demo phase ($\sim$5 minutes), the experimenter gave a paper sheet, which served as a memory aid, to the participant to write down the names of the objects taught in the current session. The paper sheet was kept by the experimenter and handed to the participants at the start of each session. This way the participants could remember the object names when they needed the robot to find these objects in the next sessions\footnote{Note that in real-world situations participants will not need to write down the names on a sheet as they will be interacting with the objects multiple times, but in our setup, it was not possible to interact with the objects long enough to remember their names from one session to the next. Sessions were typically a few days apart. On average, the time elapsed between the first and the last session was 25.3 days, with a maximum of 52 days. Further, the average time between any two consecutive sessions was 6.2 days, with a maximum of 29 days.}. The experimenter then took the tablet from the participant and loaded the program for the actual session on the tablet. The experimenter handed the tablet back to the participant and placed five objects to be taught in the session on one side of the table. The experimenter then mentioned to the participant that they can start their session and start teaching the five objects. 

The experimenter then went to a secluded area and the participant started teaching the five objects to the robot. Once the participant finished teaching, they moved to the testing phase. During the testing phase, they asked the robot to find the objects taught in the current session and the previous sessions. The participant could switch back to the teaching phase and reteach objects misclassified by the robot, as many times as they desired. After the final testing phase was finished, the experimenter came out of the secluded area and stated, “Thank you for coming today. We have a few questions about your experience today. Could you please answer them on this tablet?” The experimenter gave a different tablet to the participant to answer  questionnaires in Qualtrics \cite{qualtrics} format. After finishing the questionnaire, the experimenter thanked the participant. The participant then scheduled their next session.

In the next four sessions (each $\sim$20-30 minutes), the same procedure was repeated, except for changing the objects to be taught in each session. Figure \ref{fig:objects} shows the 25 objects used in our study.
Participants were also told that they can bring a maximum of two objects (per session) of their own choice in sessions 3-5 to teach to the robot. If participants brought their own objects, we replaced some of the objects from our set (Figure \ref{fig:objects}) with participants' objects (the total number of objects taught over 5 sessions remained 25). Further, participants did not go through a demo interaction in the next four sessions. At the end of the last session, the experimenter asked the participant to have a short interview to answer some questions describing their experience with the robot. This interview was audio recorded. Participants were remunerated \$30 CAD if they participated in all five sessions. Otherwise, if they did not complete all five sessions, they were remunerated \$6 CAD/session. Analysis of the audio data collected during the interviews is beyond the scope of this paper and will be reported in future work.

Examples of the teaching and testing phases are shown in the supplementary video. Our code is available at \url{https://github.com/aliayub7/cl_hri}.

\subsection{Measures}
\label{sec:measures}
To verify the hypotheses and thus evaluate the different learning models, we applied a range of quantitative measures in both experimental conditions.

\textbf{Subjective Measures.} After each trial, we asked participants to fill in the following questionnaire scales as subjective measurements aimed to test the hypotheses. We measured people's trust in the robot using the cognition-based trust subscale of Madsen's \textit{Human-Computer Trust (HCT)} questionnaire~\cite{Madsen00HCT} to address 
\textbf{H1.1}, \textbf{H2.1} and \textbf{H2.2}. The scale contains six individual questions that can be rated on a 5-point Likert scale, ranging from ``Strongly disagree'' to ``Strongly agree''.
We further used the \textit{Robot Social Attributes Scale (RoSAS)}~\cite{Carpinella17RoSAS} to measure how people rate the robot's social attributes to be able to accept or reject 
\textbf{H1.2}, 
\textbf{H2.3} and \textbf{H2.4}. The scale asks participants how closely they associate 18 attributes with the robot, using a 
Likert scale ranging from 1 to 7. A combination of these items forms three principal factors ``warmth'', ``competence'', and ``discomfort''. Additionally, we used the \textit{Nasa-Task Load Index (NASA-TLX)}~\cite{Hart06TLX} to estimate participants' mental workload to gain insights about 
\textbf{H1.3}, \textbf{H2.5}, and \textbf{H2.6}. \textit{TLX} is comprised of six questions that participants rate on a 21-point scale, ranging from ``Very low'' to ``Very high'', resulting in a single factor. Finally, we estimated an overall usability score using the \textit{System Usability Scale (SUS)}~\cite{Brooke95SUS} to address 
\textbf{H1.4}, \textbf{H2.7}, and \textbf{H2.8}. This scale is presented in ten questions on a 5-point Likert scale, ranging from ``1 - Strongly disagree'' to ``5 - Strongly agree'', to form a single factor. Note that we were only interested in long-term changes in the robot's social perception and hence we only employed RoSAS in the first and last session, while the other three questionnaires were presented in all five repeat sessions to allow for observing changes in between sessions.

\textbf{Objective Measures.} We also used an objective measure to analyze the performance of the three CL approaches and how it correlates with user perceptions of trust, social attributes, and usability of the continual learning robot. Classification accuracy per session (increment) has been commonly used in the continual learning literature \cite{Rebuffi_2017_CVPR,Tao_2020_CVPR,Ayub_2020_CVPR_Workshops} for quantifying the performance of CL models for object recognition tasks. Therefore, for each session, during the testing phase, we recorded the total number of objects tested by the participant and the total number of objects that were correctly found by the robot. Using this data, we calculated the accuracy $\mathcal{A}$ of the robot in each session as: 

\begin{equation}
    \mathcal{A} = \frac{total\ number\ of\ object\ correctly\ found\ in\ the\ session}{total\ number\ of\ objects\ tested\ in\ the\ session}
\end{equation}

We also report the average number of times each object was taught by the participants in the three conditions to determine the task load for teaching the robot. 

\section{Results}
\label{sec:results}
Visual inspection using quantile-quantile plots as well as applying Shapiro-Wilk tests for normality~\cite{shapiro1965analysis} suggest that none of the scores obtained using questionnaire scales are normally distributed, requiring non-parametric tests to test for potential differences. Consequently, we evaluated all questionnaire scales using a Wilcoxon rank sum test~\cite{Wilcoxon45Test} comparing the scores between the two models and five sessions, respectively. We also applied Bonferroni correction~\cite{armstrong2014use} with the Wilcoxon rank sum test to avoid false positives in our multiple statistical hypotheses testing. The analysis of each model's accuracy is discussed below, followed by the discussion of subjective measures from questionnaires, as described in Section~\ref{sec:measures}. For the remainder of the paper, we term the finetuning model as \textit{FT} (theoretically suffers from forgetting), the few-shot incremental learning (FSIL) model CBCL (SOTA model designed for FSIL to mitigate forgetting), as \textit{CBCL}, and the batch learning model as \textit{JT} (theoretical upper bound for continual learning that retrains on the data of previous sessions).

\begin{figure}[t]
\centering
\includegraphics[width=0.75\linewidth]{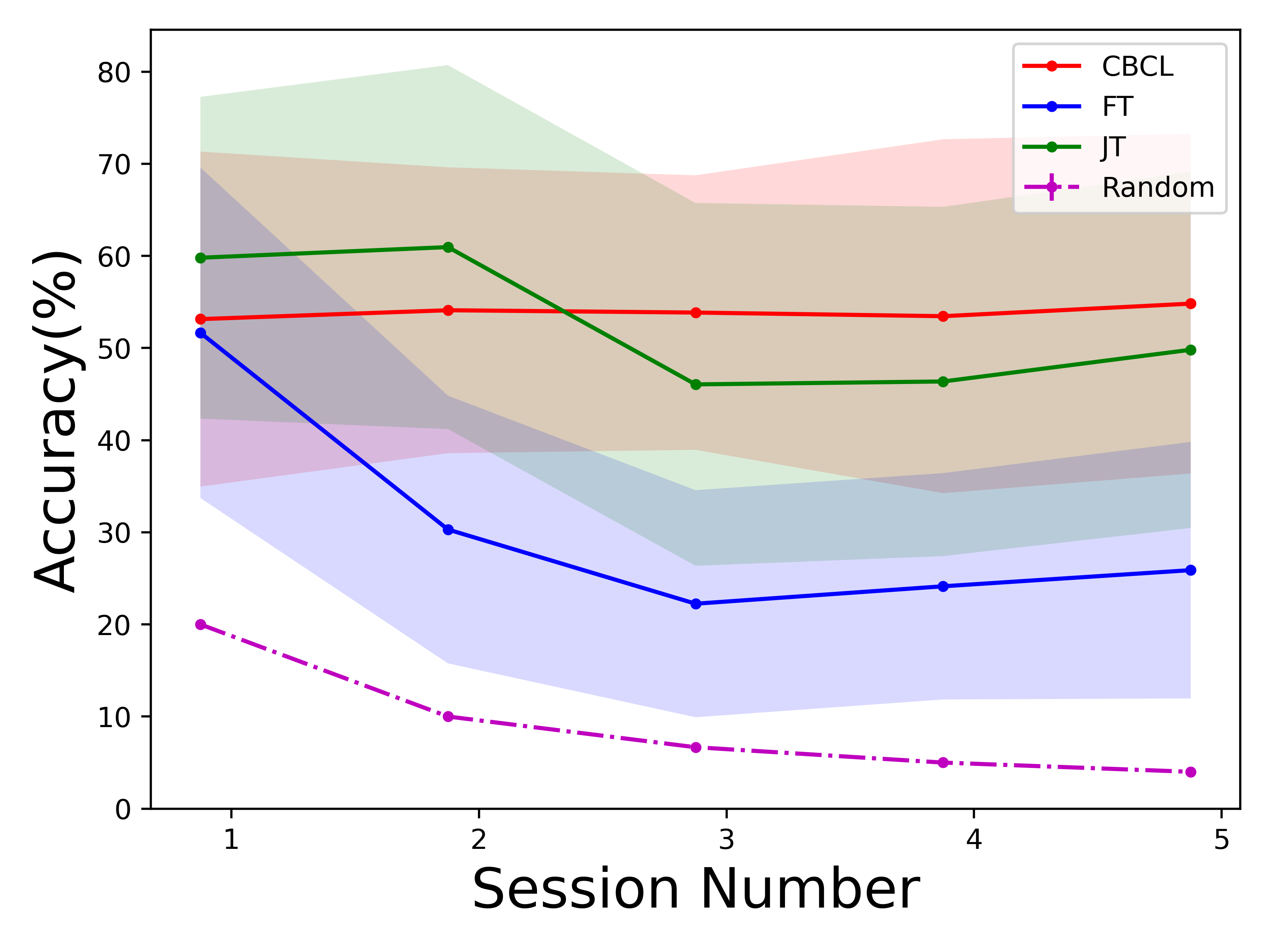}
\caption{\small Average classification accuracy of the three CL models over 5 sessions. The dotted line at the bottom represents the random chance of predicting a correct object in each session. The shaded areas represent the standard deviation for CL models.}
\label{fig:accuracy}
\end{figure}

\subsection{Classification Accuracy}
\label{sec:classification_accu}
Figure \ref{fig:accuracy} shows the classification accuracy of the three models averaged over all the participants per model. In the first session, the classification accuracy of both CBCL and FT is similar ($\mu=\sim50\%$), whereas the classification accuracy of JT is higher ($\mu=\sim60\%$). However, for the second session, FT's accuracy significantly decreased ($\mu=\sim 30\%$), and it further decreased in the next three sessions ($\mu=\sim 25\%$). CBCL's accuracy remained similar ($\mu=\sim 50\%$) in all five sessions. JT's accuracy stayed consistent in the first two sessions ($\mu=\sim 60\%$), however, it significantly decreased in the third session ($\mu=\sim 45\%$) and stayed consistent for the next three sessions. Huge variations were seen in classification accuracy for all three models in all five sessions. This variation was because of the differences in the classification accuracy of the models for different participants.

\begin{figure}[t]
    \centering
        \includegraphics[width=\linewidth,page=50,clip,trim={0cm 0cm 0cm 3.1cm}]{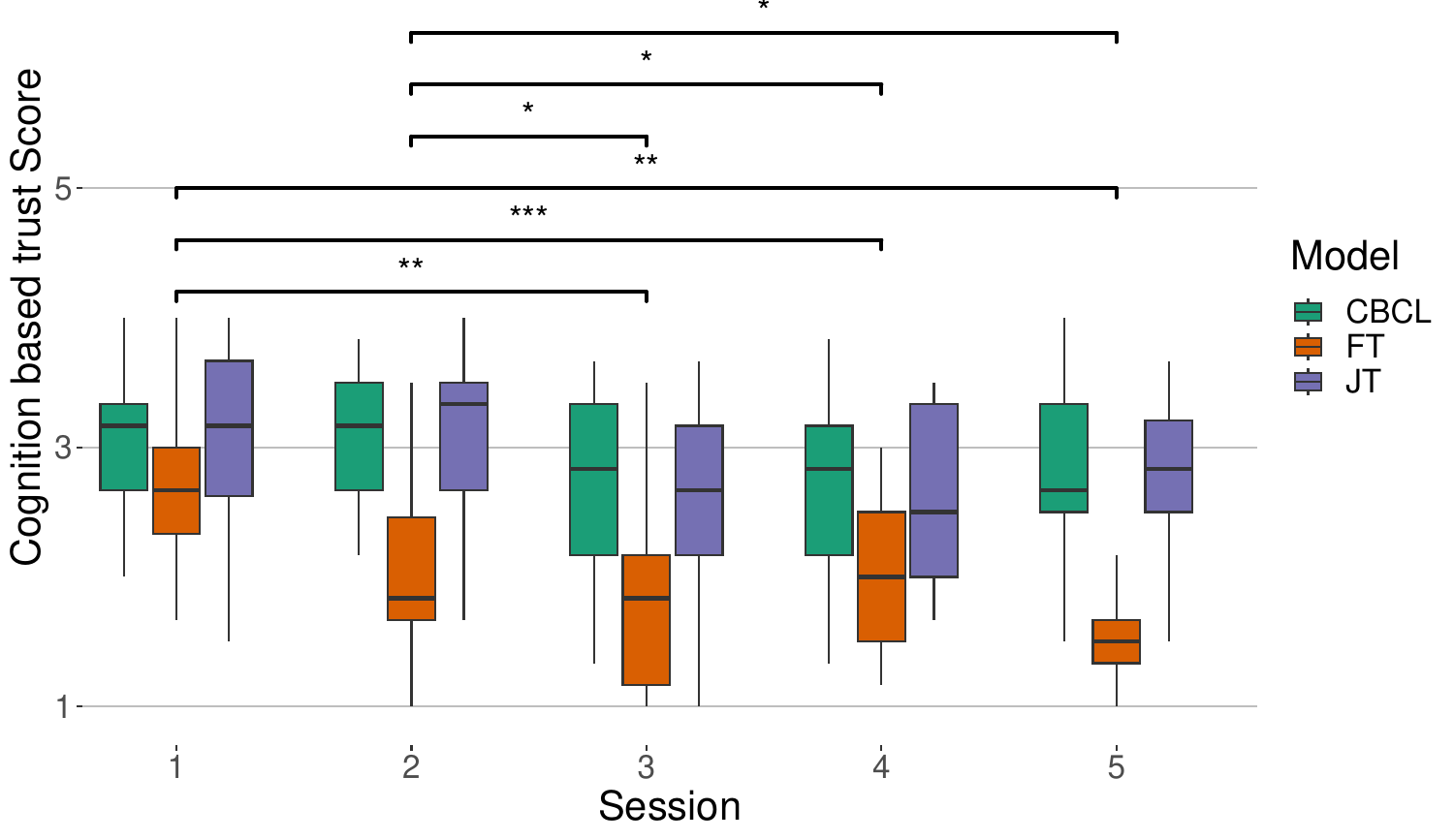}
    \caption{Boxplots for \textit{cognition based trust} scores on the \textit{HCT} scale, ranging from 1 to 5. Significance levels ($* := p < .05; ** := p < 0.01; *** :=p < 0.001$) are indicated on bars between the columns.}
    \label{fig:trust}
\end{figure}

For the \Bcondition{} we noticed that in later sessions many participants tested the robot on more new objects than old objects\footnote{Participants were not told they had to test the robot on old objects. Instead, they had flexibility regarding which objects they wanted to test in each session.} which caused the accuracy of FT for those participants to be comparable to CBCL's accuracy (notice the high standard deviation in Figure \ref{fig:accuracy}). Also, some users gave the same name to objects in different sessions. For example, some users named “green cup” in Session 1, “red cup” in Session 2, and “black mug” in Session 3, as “cup”. In such cases, FT was able to remember the previous instances of “cup” and thus the test accuracy for FT was higher.

Finally, we also calculated the average number of times each object was shown to the robot (number of training images per object) in the two conditions. Participants in all three conditions showed a similar number of images per object ($\mu=4.43$, $\sigma=3.47$ for CBCL, $\mu=5.19$, $\sigma=4.20$ for FT, and $\mu=4.82$, $\sigma=3.77$ for JT), with the highest average number of images per object in the \Bcondition{}.

\subsection{Cognition based trust}
Scores for \textit{cognition-based trust} on \textit{HCT} are calculated as mean values of six individual items with a minimum value of 1 and a maximum value of 5, resulting in an overall value of $\mu = 2.48, \sigma = 0.79$. 
Figure~\ref{fig:trust} 
details how this score differs between the subsequent experimental sessions. In particular, as displayed in Figure~\ref{fig:trust}
, trust decreases significantly only in the \Bcondition{} when comparing the first session with any of the subsequent sessions, i.e. when comparing session 1 ($\mu = 2.52, \sigma = 0.96$) to session 2 ($\mu = 1.92, \sigma = 0.79; p = 0.02584, W = 117$), to session 3 ($\mu= 1.66,\sigma = 0.81; p = 0.00329, W = 97$), to session 4 ($\mu = 1.75, \sigma = 0.88; p = 0.01015, W = 97.5$), and when comparing to session 5 ($\mu = 1.51, \sigma = 0.66; p = 0.00025, W = 69.5$). When considering the \Acondition{}, a statistically significant difference is seen only between session 1 ($\mu = 3.07, \sigma = 0.53$) and session 4 ($\mu = 2.63, \sigma = 0.67; p = 0.03689, W = 137.5$), whereas no significant differences in scores can be observed between any of the sessions for the \Ccondition{}. 

Moreover, \textit{cognition based trust} scores are significantly different between the \Acondition{} ($\mu = 2.86, \sigma = 0.65$) and \Bcondition{} ($\mu = 1.88, \sigma = 0.89$) with ($p < 0.0001, W = 1930.5$), and between the \Ccondition{} ($\mu = 2.67, \sigma = 0.87$) and \Bcondition{} with ($p<0.0001, W = 7344$), when looking at all sessions combined, and consistently across all five sessions (see Table \ref{tab:trust_detailed} for details).

\subsection{Robot social attributes}
\begin{figure}[t]
    \centering
    \begin{subfigure}{.5\linewidth}
        \includegraphics[width=\linewidth,page=6,clip,trim={0cm 0cm 0cm 1.8cm}]{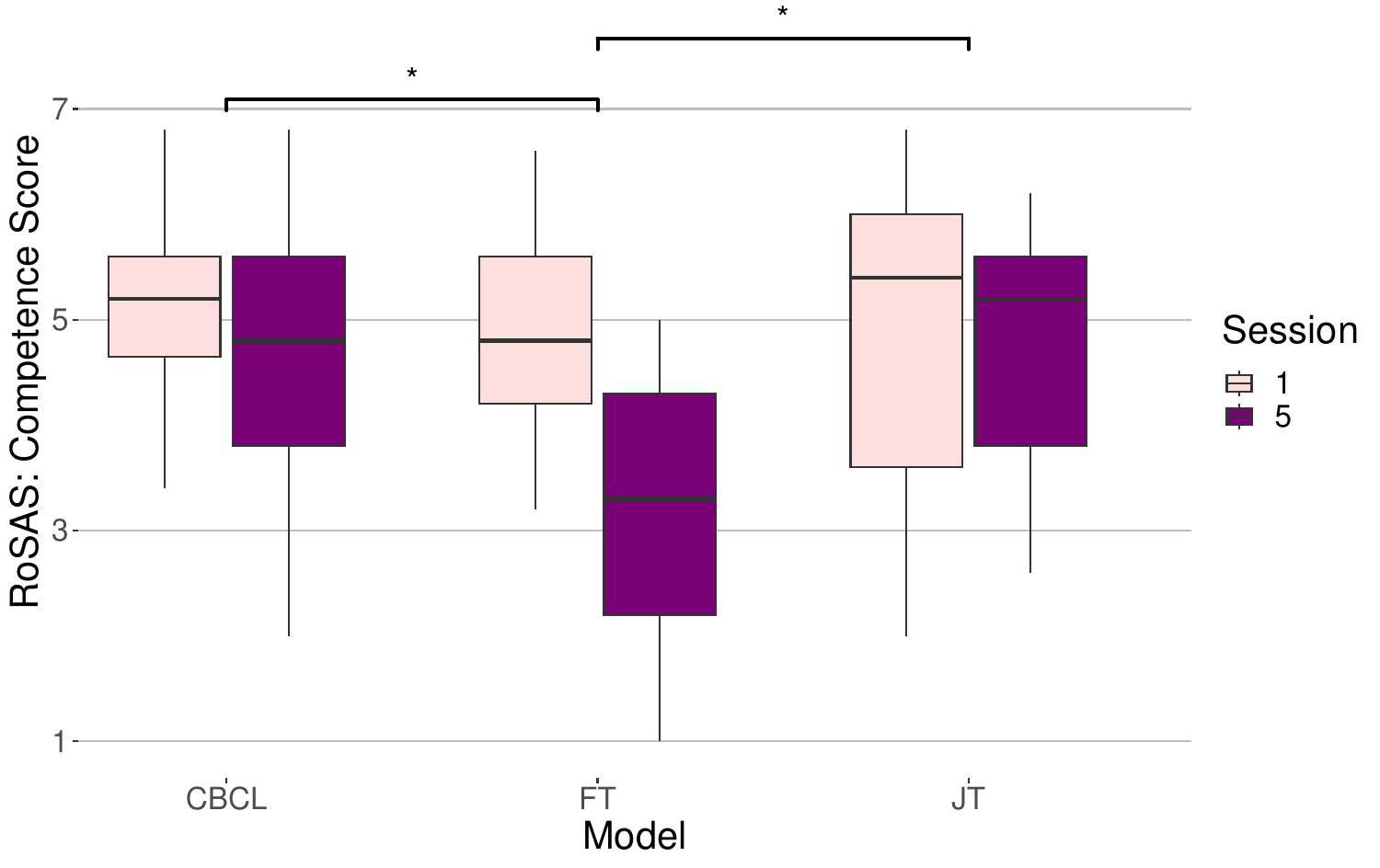}%
        \caption{RoSAS competence scores by model\\ in all sessions}
        \label{fig:rosas:model}
    \end{subfigure}%
    \begin{subfigure}{.5\linewidth}
        \includegraphics[width=\linewidth,page=8,clip,trim={0cm 0cm 0cm 1.8cm}]{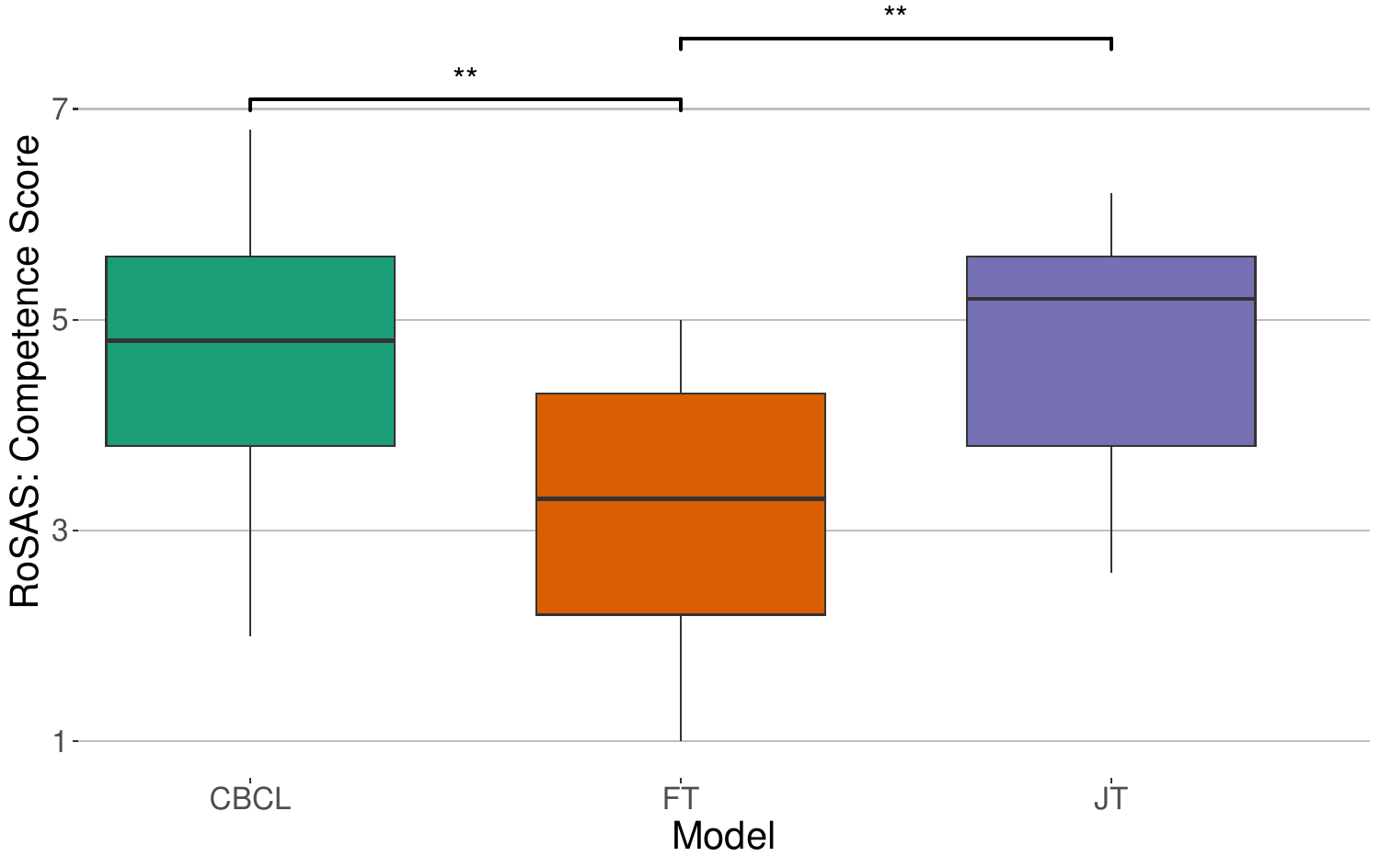}
        \caption{RoSAS competence scores by model\\ in session 5 only}
    \end{subfigure}
    \caption{Boxplots for \textit{competence} scores on the \textit{RoSAS} scale, ranging from 1 to 7. Significance levels ($* := p < .05; ** := p < 0.01; *** :=p < 0.001$) are indicated on bars between columns.}
    \label{fig:rosas}
\end{figure}

Overall scores on \textit{RoSAS} are calculated by averaging across individual items that belong to one of the subscales, ranging from 1 to 7. Resulting attribute scores are \textit{warmth}: $\mu = 2.82, \sigma = 1.36$, \textit{competence}: $\mu = 4.54, \sigma = 1.35$, and \textit{discomfort}: $\mu = 1.93, \sigma = 0.89$.
No significant differences with regard to \textit{warmth} or \textit{discomfort} can be observed when comparing experimental conditions or experimental sessions (Table \ref{tab:rosas_detailed}). \textit{Competence} scores (Figure~\ref{fig:rosas:model}), are significantly different between the \Acondition{} ($\mu = 4.77, \sigma = 1.29$) and the \Bcondition{} ($\mu = 4.01, \sigma = 1.47$) with $p = 0.0132, W = 604.5$, and between the \Ccondition{} ($\mu = 4.8, \sigma = 1.30$) and the \Bcondition{} with $p = 0.0155, W = 1127$ (Table \ref{tab:rosas_detailed}). The difference in the results between conditions is mainly driven by the ratings in the last session, whereas scores in the first session were not statistically distinguishable (Table \ref{tab:rosas_detailed}). In contrast, scores after the last session differ significantly ($p = 0.0065, W = 105.5$) between the \Acondition{} ($\mu = 4.49, \sigma = 1.41$) and the \Bcondition{} ($\mu = 3.27, \sigma = 1.17$), and between the \Ccondition{} ($\mu = 4.63, \sigma = 1.19$) and the \Bcondition{} with $p = 0.0013, W = 333.5$. Accordingly, scores significantly drop between sessions only in the \Bcondition{} (more details in Table \ref{tab:rosas_detailed}).

\subsection{Task load index}
Scores for the NASA task load index (TLX) are calculated as average values of six individual items (21-point scale, which is then translated into a score that ranges from 0 to 100). For all three models combined, the task load index remains at $\mu = 26.93, \sigma = 12.19$. No significant differences were seen between the three conditions overall (\Acondition{}: $\mu = 26.4, \sigma = 13.0$, \Bcondition{}: $\mu = 28.3, \sigma = 13.3$, \Ccondition{}: $\mu = 26.2, \sigma = 10.3$) or between any of the five sessions (see Table \ref{tab:task_load_detailed} for details).

\begin{figure}[t]
        \centering
    \begin{subfigure}{.5\linewidth}
        \includegraphics[width=\linewidth,page=27,clip,trim={0cm 0cm 0cm 1.8cm}]{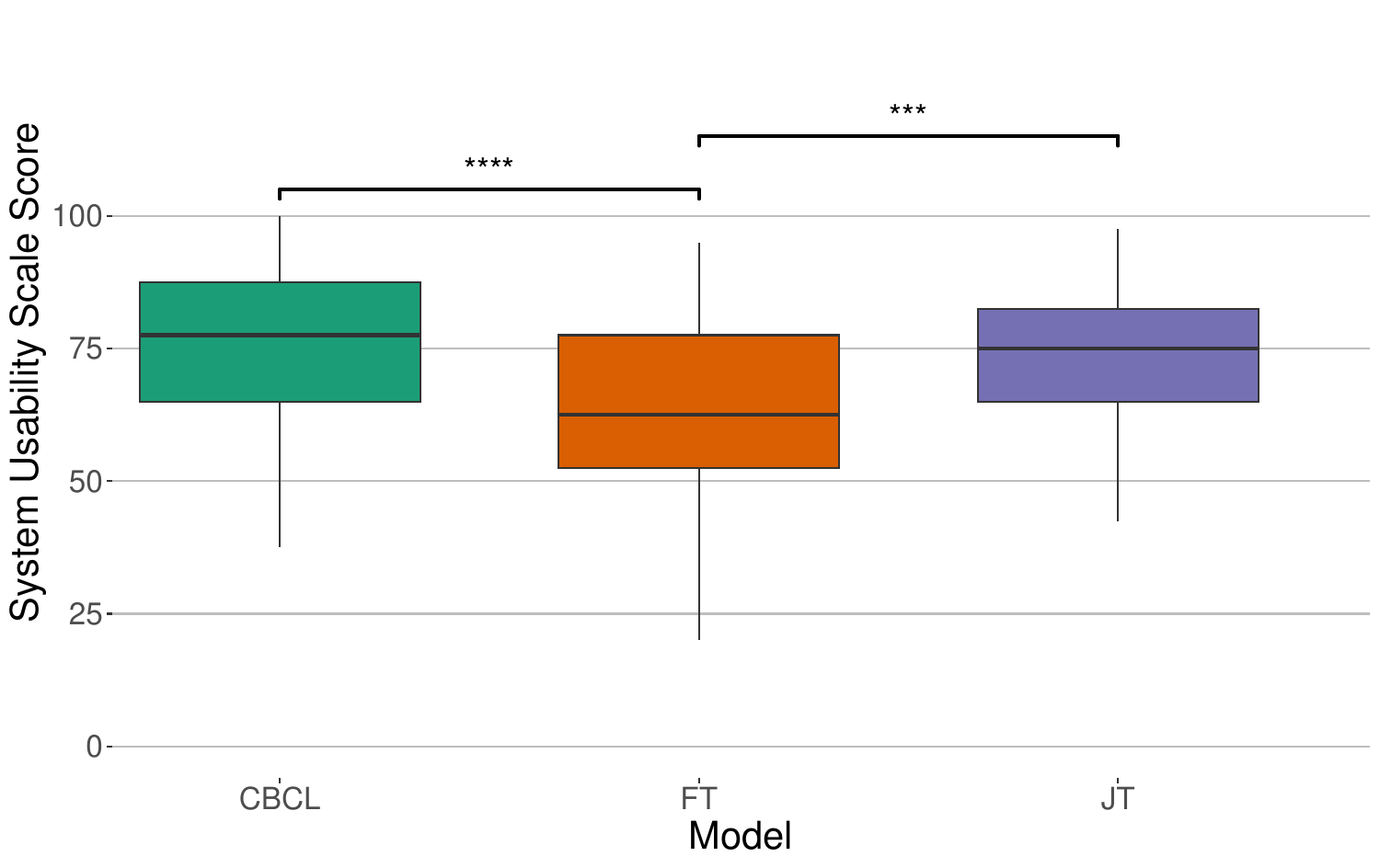}%
        \caption{Overall scores by model}
        \label{fig:sus:sessions}
    \end{subfigure}%
    \begin{subfigure}{.5\linewidth}
        \includegraphics[width=\linewidth,page=28,clip,trim={0cm 0cm 0cm 1.8cm}]{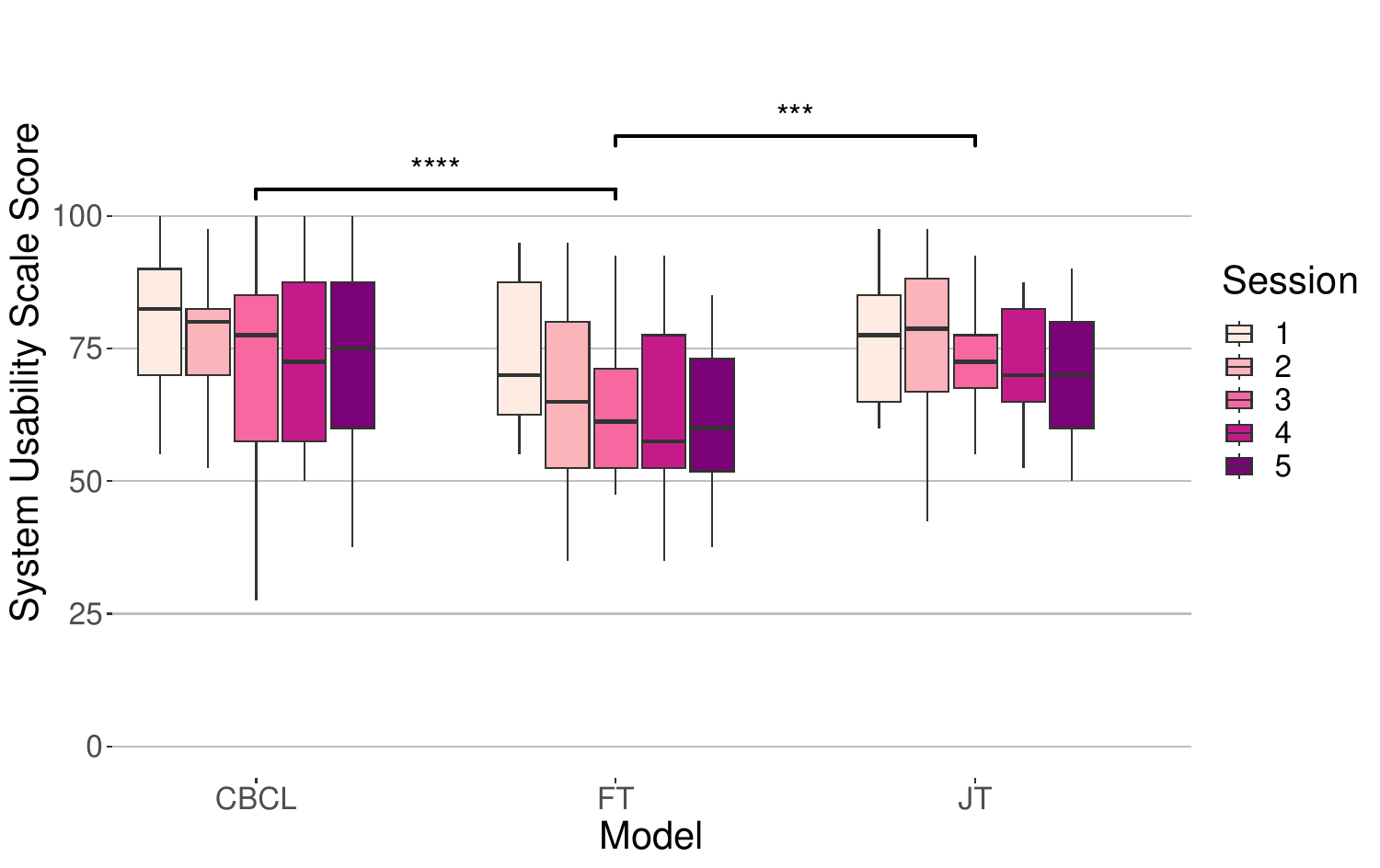}
        \caption{Scores by session for three models}
        \label{fig:sus:b}
    \end{subfigure}
    \caption{Boxplots for \textit{usability} scores on the \textit{SUS} scale, ranging from 0 to 100. Significance levels ($* := p < .05; ** := p < 0.01; *** :=p < 0.001$) are indicated on bars between columns.} 
    \label{fig:sus}
\end{figure}

\subsection{Usability}
\textit{System usability scores} are calculated as average values of ten individual questions (5-point scale, every second item inverted, which is then translated into a score that ranges from 0 to 100). The overall score is at $\mu = 70.41, \sigma = 16.29$, with significant differences between the \Acondition{} ($\mu = 74.3, \sigma = 16.6$) and the \Bcondition{} ($\mu = 63.7, \sigma = 18.8$) at $p < 0.001, W = 3441$, and between the \Ccondition{} ($\mu = 72.7, \sigma = 13.6$) and the \Bcondition{} at $p < 0.001, W = 6521$. In terms of individual sessions, the scores significantly differ between the \Acondition{} and the \Bcondition{} in sessions 2, 4, and 5, and between the \Ccondition{} and the \Bcondition{} in sessions 3, 4, and 5 (see Table \ref{tab:usability_detailed}). No statistically significant difference is seen between the \Acondition{} and the \Ccondition{} in any of the sessions. When investigating differences between the sessions, no significant change can be found in the \Acondition{} and the \Ccondition{}, whereas in the \Bcondition{}, scores in session 1 ($\mu = 73.3, \sigma = 14.0$) are significantly higher than scores in session 3 ($\mu = 62.6, \sigma = 15.9; p = 0.039, W = 130.5$), session 3 ($\mu = 58.2, \sigma = 22.3; p = 0.023, W = 108$), and session 5 ($\mu = 59.0, \sigma = 17.4; p = 0.007, W = 107$). 
No statistically significant difference was seen between sessions 1 and 2. 

\section{Discussion}
\label{sec:discussion}
Results obtained in the repeated measures experiment with the interactive system allow us to validate the hypotheses introduced in Section~\ref{sec:method} and conclusions to be drawn with regards to the research questions (\textbf{RQ1} How do human perceptions of trust, social attributes, task load, and usability evolve when interacting with a continual learning robot over multiple sessions? \textbf{RQ2} Is there a difference in participants’ perceptions of trust, social attributes, task load, and usability of a continual learning robot for different continual learning models?)
. As a general observation, results seem to be influenced by 
the change over the course of repeated sessions, how participants interacted with the models, and if the model forgot previous objects.

In comparison to other studies \cite{robinette17}, overall \textit{cognition-based trust} is rated at mediocre levels only. Such a result is within our expectations for CBCL and FT as forgetting plays an important role and the cognitive function of the system is therefore not reliably identifiable by the user. However, for JT the result was surprising as this model theoretically should not suffer from any forgetting. One reason for mediocre trust towards JT can be that JT was originally designed to learn from a large number of training samples, whereas participants only showed a few images per object (Section \ref{sec:classification_accu}) to the robot. Further, over the five sessions, the robot learned an incrementally larger number of objects for both CBCL and JT conditions, however, unlike prior work~\cite{chi_hri_23}, this did not have an accumulation effect on the perceptions of trust toward the robot. We believe that the imperfect nature of object teaching might have influenced the user's impression of the system because even the CBCL and JT approaches achieved only $\sim45-60\%$ classification accuracy in all sessions. Considering the three conditions, trust towards the system is lower in the \Bcondition{} as opposed to the \textit{CBCL} and \textit{JT} conditions, where it remains on similar levels, except between sessions 1 and 3 for the \Acondition{} where we saw a statistically significant drop in trust. 
This indicates that people, over time, lose trust in a model that forgets learned objects but they keep a similar amount of trust if it remembers previous objects. As a consequence, we can support 
\textbf{H1.1} (\textit{Users' perceptions of trust decrease in the robot over multiple sessions regardless of the CL model}) but we only find support for 
\textbf{H2.1} (\textit{A robot that forgets is perceived as less trustworthy than a robot that remembers most previously learned objects}) in the \Bcondition{} and for one session in the \Acondition{}. Hence 
\textbf{H2.1} can only be supported partially by our data. This result is consistent with the experiment's objective measures since trust seems to correlate with the classification performance of both models. The classification accuracy for \Bcondition{} decreased because of forgetting and so did the trust. For \Acondition{}, both the trust and the accuracy levels stayed similar, although accuracy remains $\sim$50\% whilst \textit{cognition based trust} decreases slightly during the course of the experiment with a significant decrease in session 3. For \Ccondition{}, accuracy slightly decreased in the final three sessions, and so did the trust in the robot. Finally, our data support \textbf{H2.2} (\textit{A robot that retrains on all previous objects is perceived as more trustworthy than a robot that does not retrain on all previous objects}) partially, as there is a statistically significant difference in perceptions of trust between the \Bcondition{} and the other two conditions, but there is no significant difference observed between \Acondition{} and \Ccondition{}. These results are promising indicating that users trust SOTA CL models, such as CBCL, that do not store and retrain on previous data similar to the theoretical upper bound JT.

For the robot's social attributes, warmth and discomfort scores 
stay reasonably low in all repeated sessions in a functional scenario with no extra social cues added to the robot, and where the interaction with the robot happens indirectly through the medium of a screen. The model choice (i.e. condition) also does not influence the experienced discomfort or warmth of the robot, making all equally good choices in terms of users' perceptions of these social attributes. This result was encouraging, showing that even a forgetful model caused little discomfort to the users. However, \textit{competence} is perceived as significantly lower by the participants after interacting multiple times with a robot that forgets (\Bcondition{}). In contrast, when using the CBCL and JT approaches, competence is rated similarly as in the first session. Therefore, all three hypotheses 
\textbf{H1.2} (\textit{Users' perception of the social attributes of the robot decreases over multiple sessions regardless of the CL model}), \textbf{H2.3} (\textit{The social attributes of a robot that forgets are perceived to be worse than those of a robot that remembers most previous objects}), and \textbf{H2.4} (\textit{The social attributes of a robot that retrains on all previous objects are perceived to be better than those of a robot that does not retrain on all previous objects}) can be supported partially by our data. For 
\textbf{H2.3}, a significant difference was seen only for competence, but not for warmth and discomfort between the three conditions and over multiple sessions. Results for discomfort are interesting because they indicate that users feel little discomfort interacting with a continual learning robot even if the robot's performance decreases over multiple sessions. Thus, only the results for competence are supported by the objective measure (classification accuracy) of the experiment. Further, for \textbf{H2.4}, similar to the results for trust, there was no difference for all three social attributes between CBCL and the theoretical upper bound JT. All three models had similarly low task load ratings, which is expected for \Bcondition{} as the model is simple and continues to forget previous objects. However, even for more complex models that mitigate forgetting, participants' workload did not increase. Neither the accuracy of the model nor any subsequent iterations affect the task load and hence 
\textbf{H2.5} (\textit{The task load for teaching and testing a robot that forgets is less than a robot that remembers most previous objects}) cannot be supported by our data. Similarly, both \textbf{H1.3} (\textit{Users' perception of the task load for teaching the robot increases over multiple sessions regardless of the CL model}) and \textbf{H2.6} (\textit{The task load for teaching and testing a robot that retrains on all previous objects is more than a robot that does not retrain on all previous objects}) are not supported since we cannot find evidence that would support any difference between the conditions with regard to task load. There is no correlation between the task load and the model's performance and the model choice. However, task load seems to be linked with the total number of images shown per object, as participants for all three models showed only a few images per object. These results are quite promising as they indicate the feasibility of personalized continual learning robots that directly learn from their users. The results also suggest that researchers might need to focus more on the task (and task load) than the choice of the ML model alone when developing continual learning robots.

The experiment results suggest an effect of model choice on the system's \textit{usability} in most sessions after the first sessions 
and therefore both 
\textbf{H1.4} (\textit{Users' perceptions of the usability of the robot decrease over multiple sessions, regardless of the CL model}) and \textbf{H2.8} (\textit{Users perceive a robot that retrains on all previous objects to be more useful and easier to use than a robot that does not retrain on all previous objects}) are supported partially. While usability scores are similar among the three approaches in the first session and partially in the second and third sessions, they drop significantly in sessions 3, 4, and 5 only when the robot uses the FT approach and is thus forgetting more frequently. The robot, however, is perceived to be equally usable over repeated sessions with the CBCL and the JT approaches. Therefore, \textbf{H2.7} (\textit{Users perceive a robot that remembers most previous objects to be more useful and easier to use than a robot that forgets}) can only be supported partially. This result is particularly interesting since \textit{usability} is not directly linked to the classification accuracy of the model. This result also shows that users find continual learning robots to be useful even when the underlying model might forget previously learned objects. This could be because some users might only care about the robot's performance on the new objects, as observed for some users in our study that did not test the FT model on many old objects in the later sessions. This might also explain the high variance in accuracy seen for the two models for different participants. Therefore, CL researchers might need to not only focus on developing optimal models but also focus on the needs and tendencies of the participants when designing continual learning robots.

Finally, we observed that JT's classification accuracy dropped by $\sim15\%$ after the first two sessions, whereas CBCL's accuracy remained consistent over all sessions. This was surprising as JT is a theoretical upper bound and trains on all the data from previous sessions, whereas CBCL only uses the data from the current session. This could be because JT was originally designed to learn from a large number of training images per object class, whereas CBCL can learn from a few images per object class. Note that in the fifth session, the time required for JT to learn new objects was $\sim20-30$ seconds because it had to retrain on the data on all the objects from the previous four sessions, whereas CBCL required $<1$ second to learn new objects even in the fifth session. This was a promising result indicating that SOTA CL models that require much less time to learn new objects can perform similarly to the theoretical upper bound for continual learning when applied to real robots interacting with real users.
However, we observed that the classification accuracy of both CBCL and JT was much lower ($\sim45-60\%$) than when tested on static datasets or with the experimenters ($>90\%$)~\cite{Rebuffi_2017_CVPR,Ayub_IROS_20}. These results indicate that the performance of the continual learning robots is quite different in the real world and it is drastically affected by the teaching style of their users. 

\section{Conclusions}
\label{sec:conclusion}
In this work, we designed a long-term user study to understand human perceptions of a continual learning robot while teaching and testing the robot over five sessions. We conducted a between-participant study with three CL models and asked participants about their perceptions of the robot in terms of trust, social attributes, task load, and usability of the system, after directly teaching and testing the robot over five sessions. Our results indicate that users' perceptions of trust, competence, and usability of the robot are negatively affected by forgetting of the CL models. Our results also indicate that the performance of even the SOTA CL models is unreliable (only $\sim$50\% accuracy) when learning from the users instead of learning on static datasets. Therefore, with the current SOTA CL models, continual learning robots are not perceived to be very trustworthy or competent by their users. However, an encouraging result was that the performance of the SOTA CL models is comparable to the theoretical upper bound for continual learning which takes a much longer time to learn new objects. Furthermore, the task load for teaching and testing the continual learning robot, and perceptions of warmth and discomfort stayed low and were not affected by the choice of the CL model. These results are encouraging as they indicate the potential feasibility of personalized continual learning robots that might learn from their users over a long period of time. Our results also indicate that future continual learning research should also focus on the task load and the needs and tendencies of the users when designing CL models that learn through human interactions.

Our user study is the first step toward testing machine learning-based CL models in the realm of HRI. We hope that these results can help machine learning and HRI researchers design CL models while considering the perceptions of human users who might interact with these systems over a long period of time. Particularly, researchers need to focus on improving the performance of the CL models when learning from human users, which might also improve users' perceptions of trust, competence, and usability of the continual learning robots. One potential direction could be to integrate CL models with human-centered AI methods, such as interpretability, and fairness, to reduce labeling ambiguities and errors caused by robots' users, which could improve the robots' performance and users' perceptions of the robots. For example, adding simple feedback, such as the number of times an object has been taught and a baseline number of images required to teach an ML model for acceptable performance, could help improve users' perception of their teaching and expectations about the robot.

\section{Limitations and Future Work}
\label{sec:limtiations}
Although we used realistic household objects and allowed participants to bring their own objects, the study was performed in a robotics lab and not in a household environment. In the future, we plan to conduct a study in a smart home with the same robot and the same learning models to determine if the household environment affects user perceptions of the continual learning robot. Further, we did not add any social cues to the robot, such as gaze or affective expressions, which might affect users' perceptions of the robot and promote more human-robot engagement. This might even improve the performance of the model through better teaching by the users. Furthermore, the available mobile manipulator robot, Fetch, has a very `functional' appearance, compared to other highly expressive social robots. In the future, we hope to expand on this study and add social capabilities to the continual learning robot. Although we conducted the first user study with a mix of experts and non-experts 
, they were all university students between the ages of 18 and 37 years. In the future, we plan to conduct this study with older adults, who might be less familiar with robots and technology in general, to understand the effectiveness of continual learning robots for assistive applications. Additionally, the study was conducted with one particular robot and with two CL models. Expanding this work in comparative studies involving more interactive and social robots with other CL models can help us understand the larger design space of continual learning robots and users' perceptions of these robots.

\backmatter

\bmhead{Acknowledgments}
This research was undertaken, in part, thanks to funding from the Canada 150 Research Chairs Program.

\section*{Declarations}

\begin{itemize}
    \item Consent for publication: All authors whose names appear on the submission approved the version to be published.
    \item Consent to participate: Informed consent was obtained from all individual participants included in the study.
    \item Code availability: Our code is available at \url{https://github.com/aliayub7/cl_hri}.
    \item Availability of data and materials: The datasets generated during and/or analyzed during the current study are available from the corresponding author upon reasonable request.
    \item Authors' contributions: Conceptualization: Ali Ayub, Patrick, Chrystopher L. Nehaniv, Kerstin Dautenhahn; Methodology: Ali Ayub, Patrick, Chrystopher L. Nehaniv, Kerstin Dautenhahn; Formal analysis and investigation: Ali Ayub, Zachary De Francesco; Writing - original draft preparation: Ali Ayub; Writing - review and editing: Ali Ayub, Patrick, Chrystopher L. Nehaniv, Kerstin Dautenhahn; Funding acquisition: Kerstin Dautenhahn; Resources: Kerstin Dautenhahn; Supervision: Chrystopher L. Nehaniv, Kerstin Dautenhahn. 
\end{itemize}

\section*{Compliance with Ethical Standards}
\begin{itemize}
    \item Funding: This research was undertaken, in part, thanks to funding from the Canada 150 Research Chairs Program.
    \item Conflict of Interest: The authors have associations or collaborations with the following domains: uwaterloo.ca, psu.edu, herts.ac.uk. The authors declare that they have no other conflicts of interest.
\end{itemize}

\bibliography{main}


\begin{thebibliography}{87}
\ifx \bisbn   \undefined \def \bisbn  #1{ISBN #1}\fi
\ifx \binits  \undefined \def \binits#1{#1}\fi
\ifx \bauthor  \undefined \def \bauthor#1{#1}\fi
\ifx \batitle  \undefined \def \batitle#1{#1}\fi
\ifx \bjtitle  \undefined \def \bjtitle#1{#1}\fi
\ifx \bvolume  \undefined \def \bvolume#1{\textbf{#1}}\fi
\ifx \byear  \undefined \def \byear#1{#1}\fi
\ifx \bissue  \undefined \def \bissue#1{#1}\fi
\ifx \bfpage  \undefined \def \bfpage#1{#1}\fi
\ifx \blpage  \undefined \def \blpage #1{#1}\fi
\ifx \burl  \undefined \def \burl#1{\textsf{#1}}\fi
\ifx \doiurl  \undefined \def \doiurl#1{\url{https://doi.org/#1}}\fi
\ifx \betal  \undefined \def \betal{\textit{et al.}}\fi
\ifx \binstitute  \undefined \def \binstitute#1{#1}\fi
\ifx \binstitutionaled  \undefined \def \binstitutionaled#1{#1}\fi
\ifx \bctitle  \undefined \def \bctitle#1{#1}\fi
\ifx \beditor  \undefined \def \beditor#1{#1}\fi
\ifx \bpublisher  \undefined \def \bpublisher#1{#1}\fi
\ifx \bbtitle  \undefined \def \bbtitle#1{#1}\fi
\ifx \bedition  \undefined \def \bedition#1{#1}\fi
\ifx \bseriesno  \undefined \def \bseriesno#1{#1}\fi
\ifx \blocation  \undefined \def \blocation#1{#1}\fi
\ifx \bsertitle  \undefined \def \bsertitle#1{#1}\fi
\ifx \bsnm \undefined \def \bsnm#1{#1}\fi
\ifx \bsuffix \undefined \def \bsuffix#1{#1}\fi
\ifx \bparticle \undefined \def \bparticle#1{#1}\fi
\ifx \barticle \undefined \def \barticle#1{#1}\fi
\bibcommenthead
\ifx \bconfdate \undefined \def \bconfdate #1{#1}\fi
\ifx \botherref \undefined \def \botherref #1{#1}\fi
\ifx \url \undefined \def \url#1{\textsf{#1}}\fi
\ifx \bchapter \undefined \def \bchapter#1{#1}\fi
\ifx \bbook \undefined \def \bbook#1{#1}\fi
\ifx \bcomment \undefined \def \bcomment#1{#1}\fi
\ifx \oauthor \undefined \def \oauthor#1{#1}\fi
\ifx \citeauthoryear \undefined \def \citeauthoryear#1{#1}\fi
\ifx \endbibitem  \undefined \def \endbibitem {}\fi
\ifx \bconflocation  \undefined \def \bconflocation#1{#1}\fi
\ifx \arxivurl  \undefined \def \arxivurl#1{\textsf{#1}}\fi
\csname PreBibitemsHook\endcsname

\bibitem{Matari17}
\begin{botherref}
\oauthor{\bsnm{Matari{\'c}}, \binits{M.J.}}:
Socially assistive robotics: Human augmentation versus automation.
Science Robotics
\textbf{2}(4)
(2017)
\end{botherref}
\endbibitem

\bibitem{petrecca_how_2018}
\begin{botherref}
\oauthor{\bsnm{Petrecca}, \binits{L.}}:
How {Robot} {Caregivers} {Will} {Help} an {Aging} {U}.{S}. {Population}
(2018).
\url{https://www.aarp.org/caregiving/home-care/info-2018/new-wave-of-caregiving-technology.html}
Accessed 2020-10-05
\end{botherref}
\endbibitem

\bibitem{Saunders16}
\begin{barticle}
\bauthor{\bsnm{Saunders}, \binits{J.}},
\bauthor{\bsnm{Syrdal}, \binits{D.S.}},
\bauthor{\bsnm{Koay}, \binits{K.L.}},
\bauthor{\bsnm{Burke}, \binits{N.}},
\bauthor{\bsnm{Dautenhahn}, \binits{K.}}:
\batitle{“{T}each {M}e–{S}how {Me}”—{E}nd-user personalization of a smart home and companion robot}.
\bjtitle{IEEE Transactions on Human-Machine Systems}
\bvolume{46}(\bissue{1}),
\bfpage{27}--\blpage{40}
(\byear{2016})
\end{barticle}
\endbibitem

\bibitem{Koay21}
\begin{barticle}
\bauthor{\bsnm{Koay}, \binits{K.L.}},
\bauthor{\bsnm{Webster}, \binits{M.}},
\bauthor{\bsnm{Dixon}, \binits{C.}},
\bauthor{\bsnm{Gainer}, \binits{P.}},
\bauthor{\bsnm{Syrdal}, \binits{D.}},
\bauthor{\bsnm{Fisher}, \binits{M.}},
\bauthor{\bsnm{Dautenhahn}, \binits{K.}}:
\batitle{Use and usability of software verification methods to detect behaviour interference when teaching an assistive home companion robot: A proof-of-concept study}.
\bjtitle{Paladyn, Journal of Behavioral Robotics}
\bvolume{12}(\bissue{1}),
\bfpage{402}--\blpage{422}
(\byear{2021})
\end{barticle}
\endbibitem

\bibitem{reiser13}
\begin{botherref}
\oauthor{\bsnm{Reiser}, \binits{U.}},
\oauthor{\bsnm{Jacobs}, \binits{T.}},
\oauthor{\bsnm{Arbeiter}, \binits{G.}},
\oauthor{\bsnm{Parlitz}, \binits{C.}},
\oauthor{\bsnm{Dautenhahn}, \binits{K.}}:
Care-o-bot$^{\mbox{\tiny ®}}$ 3 – vision of a robot butler.
Your Virtual Butler,
97--116
(2013)
\end{botherref}
\endbibitem

\bibitem{shah_hri_23}
\begin{bchapter}
\bauthor{\bsnm{Shah}, \binits{J.}},
\bauthor{\bsnm{Ayub}, \binits{A.}},
\bauthor{\bsnm{Nehaniv}, \binits{C.L.}},
\bauthor{\bsnm{Dautenhahn}, \binits{K.}}:
\bctitle{Where is my phone? towards developing an episodic memory model for companion robots to track users' salient objects}.
In: \bbtitle{Companion of the 2023 ACM/IEEE International Conference on Human-Robot Interaction}.
\bsertitle{HRI '23},
pp. \bfpage{621}--\blpage{624}.
\bpublisher{Association for Computing Machinery},
\blocation{New York, NY, USA}
(\byear{2023}).
\doiurl{10.1145/3568294.3580160}
\end{bchapter}
\endbibitem

\bibitem{Dehghan19}
\begin{bchapter}
\bauthor{\bsnm{Dehghan}, \binits{M.}},
\bauthor{\bsnm{Zhang}, \binits{Z.}},
\bauthor{\bsnm{Siam}, \binits{M.}},
\bauthor{\bsnm{Jin}, \binits{J.}},
\bauthor{\bsnm{Petrich}, \binits{L.}},
\bauthor{\bsnm{Jagersand}, \binits{M.}}:
\bctitle{Online object and task learning via human robot interaction}.
In: \bbtitle{2019 International Conference on Robotics and Automation (ICRA)},
pp. \bfpage{2132}--\blpage{2138}
(\byear{2019})
\end{bchapter}
\endbibitem

\bibitem{Valipour17}
\begin{botherref}
\oauthor{\bsnm{Valipour}, \binits{S.}},
\oauthor{\bsnm{Quintero}, \binits{C.P.}},
\oauthor{\bsnm{J{\"a}gersand}, \binits{M.}}:
Incremental learning for robot perception through hri.
2017 IEEE/RSJ International Conference on Intelligent Robots and Systems (IROS),
2772--2777
(2017)
\end{botherref}
\endbibitem

\bibitem{Ayub_IROS_20}
\begin{botherref}
\oauthor{\bsnm{Ayub}, \binits{A.}},
\oauthor{\bsnm{Wagner}, \binits{A.R.}}:
Tell me what this is: Few-shot incremental object learning by a robot.
IEEE/RSJ International Conference on Intelligent Robots and Systems (IROS)
(2020)
\end{botherref}
\endbibitem

\bibitem{thomaz09}
\begin{bchapter}
\bauthor{\bsnm{Thomaz}, \binits{A.L.}},
\bauthor{\bsnm{Cakmak}, \binits{M.}}:
\bctitle{Learning about objects with human teachers}.
In: \bbtitle{2009 4th ACM/IEEE International Conference on Human-Robot Interaction (HRI)},
pp. \bfpage{15}--\blpage{22}
(\byear{2009})
\end{bchapter}
\endbibitem

\bibitem{He_2016_CVPR}
\begin{bchapter}
\bauthor{\bsnm{He}, \binits{K.}},
\bauthor{\bsnm{Zhang}, \binits{X.}},
\bauthor{\bsnm{Ren}, \binits{S.}},
\bauthor{\bsnm{Sun}, \binits{J.}}:
\bctitle{Deep residual learning for image recognition}.
In: \bbtitle{The IEEE Conference on Computer Vision and Pattern Recognition (CVPR)}
(\byear{2016})
\end{bchapter}
\endbibitem

\bibitem{Simonyan14}
\begin{bchapter}
\bauthor{\bsnm{Simonyan}, \binits{K.}},
\bauthor{\bsnm{Zisserman}, \binits{A.}}:
\bctitle{Two-stream convolutional networks for action recognition in videos}.
In: \bbtitle{Proceedings of the 27th International Conference on Neural Information Processing Systems - Volume 1}.
\bsertitle{NIPS'14},
pp. \bfpage{568}--\blpage{576}.
\bpublisher{MIT Press},
\blocation{Cambridge, MA, USA}
(\byear{2014})
\end{bchapter}
\endbibitem

\bibitem{french19}
\begin{botherref}
\oauthor{\bsnm{French}, \binits{R.M.}}:
Dynamically constraining connectionist networks to produce distributed, orthogonal representations to reduce catastrophic interference.
Proceedings of the Sixteenth Annual Conference of the Cognitive Science Society,
335--340
(2019)
\end{botherref}
\endbibitem

\bibitem{mcclelland95}
\begin{barticle}
\bauthor{\bsnm{Mcclelland}, \binits{J.L.}},
\bauthor{\bsnm{Mcnaughton}, \binits{B.L.}},
\bauthor{\bsnm{Oreilly}, \binits{R.C.}}:
\batitle{Why there are complementary learning systems in the hippocampus and neocortex: Insights from the successes and failures of connectionist models of learning and memory.}
\bjtitle{Psychological Review}
\bvolume{102}(\bissue{3}),
\bfpage{419}--\blpage{457}
(\byear{1995}).
\doiurl{10.1037/0033-295x.102.3.419}
\end{barticle}
\endbibitem

\bibitem{Rebuffi_2017_CVPR}
\begin{bchapter}
\bauthor{\bsnm{Rebuffi}, \binits{S.-A.}},
\bauthor{\bsnm{Kolesnikov}, \binits{A.}},
\bauthor{\bsnm{Sperl}, \binits{G.}},
\bauthor{\bsnm{Lampert}, \binits{C.H.}}:
\bctitle{i{C}a{RL}: Incremental classifier and representation learning}.
In: \bbtitle{The IEEE Conference on Computer Vision and Pattern Recognition (CVPR)}
(\byear{2017})
\end{bchapter}
\endbibitem

\bibitem{costanzi21}
\begin{barticle}
\bauthor{\bsnm{Costanzi}, \binits{M.}},
\bauthor{\bsnm{Cianfanelli}, \binits{B.}},
\bauthor{\bsnm{Santirocchi}, \binits{A.}},
\bauthor{\bsnm{Lasaponara}, \binits{S.}},
\bauthor{\bsnm{Spataro}, \binits{P.}},
\bauthor{\bsnm{Rossi-Arnaud}, \binits{C.}},
\bauthor{\bsnm{Cestari}, \binits{V.}}:
\batitle{Forgetting unwanted memories: Active forgetting and implications for the development of psychological disorders}.
\bjtitle{Journal of Personalized Medicine}
\bvolume{11}(\bissue{4}),
\bfpage{241}
(\byear{2021}).
\doiurl{10.3390/jpm11040241}
\end{barticle}
\endbibitem

\bibitem{Mack18}
\begin{barticle}
\bauthor{\bsnm{Mack}, \binits{M.L.}},
\bauthor{\bsnm{Love}, \binits{B.C.}},
\bauthor{\bsnm{Preston}, \binits{A.R.}}:
\batitle{Building concepts one episode at a time: The hippocampus and concept formation}.
\bjtitle{Neuroscience Letters}
\bvolume{680},
\bfpage{31}--\blpage{38}
(\byear{2018})
\end{barticle}
\endbibitem

\bibitem{kirkpatrick17}
\begin{barticle}
\bauthor{\bsnm{Kirkpatrick}, \binits{J.}},
\bauthor{\bsnm{Pascanu}, \binits{R.}},
\bauthor{\bsnm{Rabinowitz}, \binits{N.C.}},
\bauthor{\bsnm{Veness}, \binits{J.}},
\bauthor{\bsnm{Desjardins}, \binits{G.}},
\bauthor{\bsnm{Rusu}, \binits{A.A.}},
\bauthor{\bsnm{Milan}, \binits{K.}},
\bauthor{\bsnm{Quan}, \binits{J.}},
\bauthor{\bsnm{Ramalho}, \binits{T.}},
\bauthor{\bsnm{Grabska-Barwinska}, \binits{A.}},
\bauthor{\bsnm{Hassabis}, \binits{D.}},
\bauthor{\bsnm{Clopath}, \binits{C.}},
\bauthor{\bsnm{Kumaran}, \binits{D.}},
\bauthor{\bsnm{Hadsell}, \binits{R.}}:
\batitle{Overcoming catastrophic forgetting in neural networks}.
\bjtitle{Proceedings of the National Academy of Sciences of the United States of America}
\bvolume{114}(\bissue{13}),
\bfpage{3521}--\blpage{3526}
(\byear{2017})
\end{barticle}
\endbibitem

\bibitem{kemker18}
\begin{bchapter}
\bauthor{\bsnm{Kemker}, \binits{R.}},
\bauthor{\bsnm{Kanan}, \binits{C.}}:
\bctitle{Fearnet: Brain-inspired model for incremental learning}.
In: \bbtitle{International Conference on Learning Representations}
(\byear{2018}).
\burl{https://openreview.net/forum?id=SJ1Xmf-Rb}
\end{bchapter}
\endbibitem

\bibitem{mudt2020}
\begin{botherref}
\oauthor{\bsnm{Mundt}, \binits{M.}},
\oauthor{\bsnm{Hong}, \binits{Y.W.}},
\oauthor{\bsnm{Pliushch}, \binits{I.}},
\oauthor{\bsnm{Ramesh}, \binits{V.}}:
A wholistic view of continual learning with deep neural networks: Forgotten lessons and the bridge to active and open world learning.
CoRR
\textbf{abs/2009.01797}
(2020)
\end{botherref}
\endbibitem

\bibitem{mundt2022clevacompass}
\begin{bchapter}
\bauthor{\bsnm{Mundt}, \binits{M.}},
\bauthor{\bsnm{Lang}, \binits{S.}},
\bauthor{\bsnm{Delfosse}, \binits{Q.}},
\bauthor{\bsnm{Kersting}, \binits{K.}}:
\bctitle{{CLEVA}-compass: A continual learning evaluation assessment compass to promote research transparency and comparability}.
In: \bbtitle{International Conference on Learning Representations}
(\byear{2022}).
\burl{https://openreview.net/forum?id=rHMaBYbkkRJ}
\end{bchapter}
\endbibitem

\bibitem{Smith_2021_ICCV}
\begin{bchapter}
\bauthor{\bsnm{Smith}, \binits{J.}},
\bauthor{\bsnm{Hsu}, \binits{Y.-C.}},
\bauthor{\bsnm{Balloch}, \binits{J.}},
\bauthor{\bsnm{Shen}, \binits{Y.}},
\bauthor{\bsnm{Jin}, \binits{H.}},
\bauthor{\bsnm{Kira}, \binits{Z.}}:
\bctitle{Always be dreaming: A new approach for data-free class-incremental learning}.
In: \bbtitle{Proceedings of the IEEE/CVF International Conference on Computer Vision (ICCV)},
pp. \bfpage{9374}--\blpage{9384}
(\byear{2021})
\end{bchapter}
\endbibitem

\bibitem{lomonaco17}
\begin{bchapter}
\bauthor{\bsnm{Lomonaco}, \binits{V.}},
\bauthor{\bsnm{Maltoni}, \binits{D.}}:
\bctitle{Core50: a new dataset and benchmark for continuous object recognition}.
In: \bbtitle{Proceedings of the 1st Annual Conference on Robot Learning},
vol. \bseriesno{78},
pp. \bfpage{17}--\blpage{26}
(\byear{2017})
\end{bchapter}
\endbibitem

\bibitem{Chaudhry_2018_ECCV}
\begin{bchapter}
\bauthor{\bsnm{Chaudhry}, \binits{A.}},
\bauthor{\bsnm{Dokania}, \binits{P.K.}},
\bauthor{\bsnm{Ajanthan}, \binits{T.}},
\bauthor{\bsnm{Torr}, \binits{P.H.S.}}:
\bctitle{Riemannian walk for incremental learning: Understanding forgetting and intransigence}.
In: \bbtitle{The European Conference on Computer Vision (ECCV)}
(\byear{2018})
\end{bchapter}
\endbibitem

\bibitem{hayes20}
\begin{bchapter}
\bauthor{\bsnm{Hayes}, \binits{T.L.}},
\bauthor{\bsnm{Kafle}, \binits{K.}},
\bauthor{\bsnm{Shrestha}, \binits{R.}},
\bauthor{\bsnm{Acharya}, \binits{M.}},
\bauthor{\bsnm{Kanan}, \binits{C.}}:
\bctitle{Remind your neural network to prevent catastrophic forgetting}.
In: \beditor{\bsnm{Vedaldi}, \binits{A.}},
\beditor{\bsnm{Bischof}, \binits{H.}},
\beditor{\bsnm{Brox}, \binits{T.}},
\beditor{\bsnm{Frahm}, \binits{J.-M.}} (eds.)
\bbtitle{Computer Vision -- ECCV 2020},
pp. \bfpage{466}--\blpage{483}.
\bpublisher{Springer},
\blocation{Cham}
(\byear{2020})
\end{bchapter}
\endbibitem

\bibitem{smith2022incremental}
\begin{bchapter}
\bauthor{\bsnm{Smith}, \binits{J.S.}},
\bauthor{\bsnm{Seymour}, \binits{Z.}},
\bauthor{\bsnm{Chiu}, \binits{H.-P.}}:
\bctitle{Incremental learning with differentiable architecture and forgetting search}.
In: \bbtitle{2022 International Joint Conference on Neural Networks (IJCNN)},
pp. \bfpage{01}--\blpage{08}
(\byear{2022}).
\bcomment{IEEE}
\end{bchapter}
\endbibitem

\bibitem{Li18}
\begin{barticle}
\bauthor{\bsnm{Li}, \binits{Z.}},
\bauthor{\bsnm{Hoiem}, \binits{D.}}:
\batitle{Learning without forgetting}.
\bjtitle{IEEE Transactions on Pattern Analysis and Machine Intelligence}
\bvolume{40}(\bissue{12}),
\bfpage{2935}--\blpage{2947}
(\byear{2018})
\end{barticle}
\endbibitem

\bibitem{ayub2021eec}
\begin{bchapter}
\bauthor{\bsnm{Ayub}, \binits{A.}},
\bauthor{\bsnm{Wagner}, \binits{A.}}:
\bctitle{Eec: Learning to encode and regenerate images for continual learning}.
In: \bbtitle{International Conference on Learning Representations}
(\byear{2021}).
\burl{https://openreview.net/forum?id=lWaz5a9lcFU}
\end{bchapter}
\endbibitem

\bibitem{Ostapenko_2019_CVPR}
\begin{bchapter}
\bauthor{\bsnm{Ostapenko}, \binits{O.}},
\bauthor{\bsnm{Puscas}, \binits{M.}},
\bauthor{\bsnm{Klein}, \binits{T.}},
\bauthor{\bsnm{Jahnichen}, \binits{P.}},
\bauthor{\bsnm{Nabi}, \binits{M.}}:
\bctitle{Learning to remember: A synaptic plasticity driven framework for continual learning}.
In: \bbtitle{The IEEE Conference on Computer Vision and Pattern Recognition (CVPR)},
pp. \bfpage{11321}--\blpage{11329}
(\byear{2019})
\end{bchapter}
\endbibitem

\bibitem{Ayub_2020_CVPR_Workshops}
\begin{bchapter}
\bauthor{\bsnm{Ayub}, \binits{A.}},
\bauthor{\bsnm{Wagner}, \binits{A.R.}}:
\bctitle{Cognitively-inspired model for incremental learning using a few examples}.
In: \bbtitle{The IEEE/CVF Conference on Computer Vision and Pattern Recognition (CVPR) Workshops}
(\byear{2020})
\end{bchapter}
\endbibitem

\bibitem{Tao_2020_CVPR}
\begin{bchapter}
\bauthor{\bsnm{Tao}, \binits{X.}},
\bauthor{\bsnm{Hong}, \binits{X.}},
\bauthor{\bsnm{Chang}, \binits{X.}},
\bauthor{\bsnm{Dong}, \binits{S.}},
\bauthor{\bsnm{Wei}, \binits{X.}},
\bauthor{\bsnm{Gong}, \binits{Y.}}:
\bctitle{Few-shot class-incremental learning}.
In: \bbtitle{Proceedings of the IEEE/CVF Conference on Computer Vision and Pattern Recognition (CVPR)}
(\byear{2020})
\end{bchapter}
\endbibitem

\bibitem{lesort2020continual}
\begin{barticle}
\bauthor{\bsnm{Lesort}, \binits{T.}},
\bauthor{\bsnm{Lomonaco}, \binits{V.}},
\bauthor{\bsnm{Stoian}, \binits{A.}},
\bauthor{\bsnm{Maltoni}, \binits{D.}},
\bauthor{\bsnm{Filliat}, \binits{D.}},
\bauthor{\bsnm{D{\'\i}az-Rodr{\'\i}guez}, \binits{N.}}:
\batitle{Continual learning for robotics: Definition, framework, learning strategies, opportunities and challenges}.
\bjtitle{Information fusion}
\bvolume{58},
\bfpage{52}--\blpage{68}
(\byear{2020})
\end{barticle}
\endbibitem

\bibitem{tao_2020_ECCV}
\begin{bchapter}
\bauthor{\bsnm{Tao}, \binits{X.}},
\bauthor{\bsnm{Chang}, \binits{X.}},
\bauthor{\bsnm{Hong}, \binits{X.}},
\bauthor{\bsnm{Wei}, \binits{X.}},
\bauthor{\bsnm{Gong}, \binits{Y.}}:
\bctitle{Topology-preserving class-incremental learning}.
In: \beditor{\bsnm{Vedaldi}, \binits{A.}},
\beditor{\bsnm{Bischof}, \binits{H.}},
\beditor{\bsnm{Brox}, \binits{T.}},
\beditor{\bsnm{Frahm}, \binits{J.-M.}} (eds.)
\bbtitle{Computer Vision -- ECCV 2020},
pp. \bfpage{254}--\blpage{270}.
\bpublisher{Springer},
\blocation{Cham}
(\byear{2020})
\end{bchapter}
\endbibitem

\bibitem{Zhang_2021_CVPR}
\begin{bchapter}
\bauthor{\bsnm{Zhang}, \binits{C.}},
\bauthor{\bsnm{Song}, \binits{N.}},
\bauthor{\bsnm{Lin}, \binits{G.}},
\bauthor{\bsnm{Zheng}, \binits{Y.}},
\bauthor{\bsnm{Pan}, \binits{P.}},
\bauthor{\bsnm{Xu}, \binits{Y.}}:
\bctitle{Few-shot incremental learning with continually evolved classifiers}.
In: \bbtitle{Proceedings of the IEEE/CVF Conference on Computer Vision and Pattern Recognition (CVPR)},
pp. \bfpage{12455}--\blpage{12464}
(\byear{2021})
\end{bchapter}
\endbibitem

\bibitem{paetzel20}
\begin{bchapter}
\bauthor{\bsnm{Paetzel}, \binits{M.}},
\bauthor{\bsnm{Perugia}, \binits{G.}},
\bauthor{\bsnm{Castellano}, \binits{G.}}:
\bctitle{The persistence of first impressions: The effect of repeated interactions on the perception of a social robot}.
In: \bbtitle{Proceedings of the 2020 ACM/IEEE International Conference on Human-Robot Interaction}.
\bsertitle{HRI '20},
pp. \bfpage{73}--\blpage{82}.
\bpublisher{Association for Computing Machinery},
\blocation{New York, NY, USA}
(\byear{2020}).
\doiurl{10.1145/3319502.3374786}.
\burl{https://doi.org/10.1145/3319502.3374786}
\end{bchapter}
\endbibitem

\bibitem{lyons2023explanations}
\begin{barticle}
\bauthor{\bsnm{Lyons}, \binits{J.B.}},
\bauthor{\bparticle{aldin} \bsnm{Hamdan}, \binits{I.}},
\bauthor{\bsnm{Vo}, \binits{T.Q.}}:
\batitle{Explanations and trust: What happens to trust when a robot partner does something unexpected?}
\bjtitle{Computers in Human Behavior}
\bvolume{138},
\bfpage{107473}
(\byear{2023})
\end{barticle}
\endbibitem

\bibitem{de2020towards}
\begin{barticle}
\bauthor{\bsnm{De~Visser}, \binits{E.J.}},
\bauthor{\bsnm{Peeters}, \binits{M.M.}},
\bauthor{\bsnm{Jung}, \binits{M.F.}},
\bauthor{\bsnm{Kohn}, \binits{S.}},
\bauthor{\bsnm{Shaw}, \binits{T.H.}},
\bauthor{\bsnm{Pak}, \binits{R.}},
\bauthor{\bsnm{Neerincx}, \binits{M.A.}}:
\batitle{Towards a theory of longitudinal trust calibration in human--robot teams}.
\bjtitle{International journal of social robotics}
\bvolume{12}(\bissue{2}),
\bfpage{459}--\blpage{478}
(\byear{2020})
\end{barticle}
\endbibitem

\bibitem{rossi17}
\begin{bchapter}
\bauthor{\bsnm{Rossi}, \binits{A.}},
\bauthor{\bsnm{Dautenhahn}, \binits{K.}},
\bauthor{\bsnm{Koay}, \binits{K.L.}},
\bauthor{\bsnm{Saunders}, \binits{J.}}:
\bctitle{Investigating human perceptions of trust in robots for safe hri in home environments}.
In: \bbtitle{Proceedings of the Companion of the 2017 ACM/IEEE International Conference on Human-Robot Interaction}.
\bsertitle{HRI '17},
pp. \bfpage{375}--\blpage{376}.
\bpublisher{Association for Computing Machinery},
\blocation{New York, NY, USA}
(\byear{2017}).
\doiurl{10.1145/3029798.3034822}.
\burl{https://doi.org/10.1145/3029798.3034822}
\end{bchapter}
\endbibitem

\bibitem{nayyar18}
\begin{bchapter}
\bauthor{\bsnm{Nayyar}, \binits{M.}},
\bauthor{\bsnm{Wagner}, \binits{A.R.}}:
\bctitle{When should a robot apologize? understanding how timing affects human-robot trust repair}.
In: \beditor{\bsnm{Ge}, \binits{S.S.}},
\beditor{\bsnm{Cabibihan}, \binits{J.-J.}},
\beditor{\bsnm{Salichs}, \binits{M.A.}},
\beditor{\bsnm{Broadbent}, \binits{E.}},
\beditor{\bsnm{He}, \binits{H.}},
\beditor{\bsnm{Wagner}, \binits{A.R.}},
\beditor{\bsnm{Castro-Gonz{\'a}lez}, \binits{{\'A}.}} (eds.)
\bbtitle{Social Robotics},
pp. \bfpage{265}--\blpage{274}.
\bpublisher{Springer},
\blocation{Cham}
(\byear{2018})
\end{bchapter}
\endbibitem

\bibitem{andras2018trusting}
\begin{barticle}
\bauthor{\bsnm{Andras}, \binits{P.}},
\bauthor{\bsnm{Esterle}, \binits{L.}},
\bauthor{\bsnm{Guckert}, \binits{M.}},
\bauthor{\bsnm{Han}, \binits{T.A.}},
\bauthor{\bsnm{Lewis}, \binits{P.R.}},
\bauthor{\bsnm{Milanovic}, \binits{K.}},
\bauthor{\bsnm{Payne}, \binits{T.}},
\bauthor{\bsnm{Perret}, \binits{C.}},
\bauthor{\bsnm{Pitt}, \binits{J.}},
\bauthor{\bsnm{Powers}, \binits{S.T.}}, \betal:
\batitle{Trusting intelligent machines: Deepening trust within socio-technical systems}.
\bjtitle{IEEE Technology and Society Magazine}
\bvolume{37}(\bissue{4}),
\bfpage{76}--\blpage{83}
(\byear{2018})
\end{barticle}
\endbibitem

\bibitem{esterwood2021you}
\begin{bchapter}
\bauthor{\bsnm{Esterwood}, \binits{C.}},
\bauthor{\bsnm{Robert}, \binits{L.P.}}:
\bctitle{Do you still trust me? human-robot trust repair strategies}.
In: \bbtitle{2021 30th IEEE International Conference on Robot \& Human Interactive Communication (RO-MAN)},
pp. \bfpage{183}--\blpage{188}
(\byear{2021}).
\bcomment{IEEE}
\end{bchapter}
\endbibitem

\bibitem{chi_hri_23}
\begin{bchapter}
\bauthor{\bsnm{Chi}, \binits{V.B.}},
\bauthor{\bsnm{Malle}, \binits{B.F.}}:
\bctitle{People dynamically update trust when interactively teaching robots}.
In: \bbtitle{Proceedings of the 2023 ACM/IEEE International Conference on Human-Robot Interaction}.
\bsertitle{HRI '23},
pp. \bfpage{554}--\blpage{564}.
\bpublisher{Association for Computing Machinery},
\blocation{New York, NY, USA}
(\byear{2023}).
\doiurl{10.1145/3568162.3576962}
\end{bchapter}
\endbibitem

\bibitem{scheunemann22}
\begin{barticle}
\bauthor{\bsnm{Scheunemann}, \binits{M.M.}},
\bauthor{\bsnm{Salge}, \binits{C.}},
\bauthor{\bsnm{Polani}, \binits{D.}},
\bauthor{\bsnm{Dautenhahn}, \binits{K.}}:
\batitle{Human perception of intrinsically motivated autonomy in human-robot interaction}.
\bjtitle{Adaptive Behavior}
\bvolume{30}(\bissue{5}),
\bfpage{451}--\blpage{472}
(\byear{2022})
\end{barticle}
\endbibitem

\bibitem{scassellati2018improving}
\begin{barticle}
\bauthor{\bsnm{Scassellati}, \binits{B.}},
\bauthor{\bsnm{Boccanfuso}, \binits{L.}},
\bauthor{\bsnm{Huang}, \binits{C.-M.}},
\bauthor{\bsnm{Mademtzi}, \binits{M.}},
\bauthor{\bsnm{Qin}, \binits{M.}},
\bauthor{\bsnm{Salomons}, \binits{N.}},
\bauthor{\bsnm{Ventola}, \binits{P.}},
\bauthor{\bsnm{Shic}, \binits{F.}}:
\batitle{Improving social skills in children with asd using a long-term, in-home social robot}.
\bjtitle{Science Robotics}
\bvolume{3}(\bissue{21}),
\bfpage{7544}
(\byear{2018})
\end{barticle}
\endbibitem

\bibitem{de2016long}
\begin{barticle}
\bauthor{\bsnm{De~Graaf}, \binits{M.M.}},
\bauthor{\bsnm{Ben~Allouch}, \binits{S.}},
\bauthor{\bparticle{van} \bsnm{Dijk}, \binits{J.A.}}:
\batitle{Long-term evaluation of a social robot in real homes}.
\bjtitle{Interaction studies}
\bvolume{17}(\bissue{3}),
\bfpage{462}--\blpage{491}
(\byear{2016})
\end{barticle}
\endbibitem

\bibitem{de2017they}
\begin{bchapter}
\bauthor{\bsnm{De~Graaf}, \binits{M.}},
\bauthor{\bsnm{Ben~Allouch}, \binits{S.}},
\bauthor{\bsnm{Van~Dijk}, \binits{J.}}:
\bctitle{Why do they refuse to use my robot? reasons for non-use derived from a long-term home study}.
In: \bbtitle{Proceedings of the 2017 ACM/IEEE International Conference on Human-Robot Interaction},
pp. \bfpage{224}--\blpage{233}
(\byear{2017})
\end{bchapter}
\endbibitem

\bibitem{kosch_2023}
\begin{barticle}
\bauthor{\bsnm{Kosch}, \binits{T.}},
\bauthor{\bsnm{Karolus}, \binits{J.}},
\bauthor{\bsnm{Zagermann}, \binits{J.}},
\bauthor{\bsnm{Reiterer}, \binits{H.}},
\bauthor{\bsnm{Schmidt}, \binits{A.}},
\bauthor{\bsnm{Wo\'{z}niak}, \binits{P.W.}}:
\batitle{A survey on measuring cognitive workload in human-computer interaction}.
\bjtitle{ACM Comput. Surv.}
(\byear{2023}).
\doiurl{10.1145/3582272}
\end{barticle}
\endbibitem

\bibitem{zhang2022trans4trans}
\begin{barticle}
\bauthor{\bsnm{Zhang}, \binits{J.}},
\bauthor{\bsnm{Yang}, \binits{K.}},
\bauthor{\bsnm{Constantinescu}, \binits{A.}},
\bauthor{\bsnm{Peng}, \binits{K.}},
\bauthor{\bsnm{M{\"u}ller}, \binits{K.}},
\bauthor{\bsnm{Stiefelhagen}, \binits{R.}}:
\batitle{Trans4trans: Efficient transformer for transparent object and semantic scene segmentation in real-world navigation assistance}.
\bjtitle{IEEE Transactions on Intelligent Transportation Systems}
\bvolume{23}(\bissue{10}),
\bfpage{19173}--\blpage{19186}
(\byear{2022})
\end{barticle}
\endbibitem

\bibitem{Wise16}
\begin{bchapter}
\bauthor{\bsnm{Wise}, \binits{M.}},
\bauthor{\bsnm{Ferguson}, \binits{M.}},
\bauthor{\bsnm{King}, \binits{D.}},
\bauthor{\bsnm{Diehr}, \binits{E.}},
\bauthor{\bsnm{Dymesich}, \binits{D.}}:
\bctitle{Fetch and freight: Standard platforms for service robot applications}.
In: \bbtitle{IJCAI, Workshop on Autonomous Mobile Service Robots}
(\byear{2016})
\end{bchapter}
\endbibitem

\bibitem{Wu_2019_CVPR}
\begin{bchapter}
\bauthor{\bsnm{Wu}, \binits{Y.}},
\bauthor{\bsnm{Chen}, \binits{Y.}},
\bauthor{\bsnm{Wang}, \binits{L.}},
\bauthor{\bsnm{Ye}, \binits{Y.}},
\bauthor{\bsnm{Liu}, \binits{Z.}},
\bauthor{\bsnm{Guo}, \binits{Y.}},
\bauthor{\bsnm{Fu}, \binits{Y.}}:
\bctitle{Large scale incremental learning}.
In: \bbtitle{The IEEE Conference on Computer Vision and Pattern Recognition (CVPR)}
(\byear{2019})
\end{bchapter}
\endbibitem

\bibitem{Castro_2018_ECCV}
\begin{bchapter}
\bauthor{\bsnm{Castro}, \binits{F.M.}},
\bauthor{\bsnm{Marin-Jimenez}, \binits{M.J.}},
\bauthor{\bsnm{Guil}, \binits{N.}},
\bauthor{\bsnm{Schmid}, \binits{C.}},
\bauthor{\bsnm{Alahari}, \binits{K.}}:
\bctitle{End-to-end incremental learning}.
In: \bbtitle{The European Conference on Computer Vision (ECCV)}
(\byear{2018})
\end{bchapter}
\endbibitem

\bibitem{Kang_2022_CVPR}
\begin{bchapter}
\bauthor{\bsnm{Kang}, \binits{M.}},
\bauthor{\bsnm{Park}, \binits{J.}},
\bauthor{\bsnm{Han}, \binits{B.}}:
\bctitle{Class-incremental learning by knowledge distillation with adaptive feature consolidation}.
In: \bbtitle{Proceedings of the IEEE/CVF Conference on Computer Vision and Pattern Recognition (CVPR)},
pp. \bfpage{16071}--\blpage{16080}
(\byear{2022})
\end{bchapter}
\endbibitem

\bibitem{Hou_2019_CVPR}
\begin{bchapter}
\bauthor{\bsnm{Hou}, \binits{S.}},
\bauthor{\bsnm{Pan}, \binits{X.}},
\bauthor{\bsnm{Loy}, \binits{C.C.}},
\bauthor{\bsnm{Wang}, \binits{Z.}},
\bauthor{\bsnm{Lin}, \binits{D.}}:
\bctitle{Learning a unified classifier incrementally via rebalancing}.
In: \bbtitle{The IEEE Conference on Computer Vision and Pattern Recognition (CVPR)}
(\byear{2019})
\end{bchapter}
\endbibitem

\bibitem{hayes2021replay}
\begin{barticle}
\bauthor{\bsnm{Hayes}, \binits{T.L.}},
\bauthor{\bsnm{Krishnan}, \binits{G.P.}},
\bauthor{\bsnm{Bazhenov}, \binits{M.}},
\bauthor{\bsnm{Siegelmann}, \binits{H.T.}},
\bauthor{\bsnm{Sejnowski}, \binits{T.J.}},
\bauthor{\bsnm{Kanan}, \binits{C.}}:
\batitle{Replay in deep learning: Current approaches and missing biological elements}.
\bjtitle{Neural computation}
\bvolume{33}(\bissue{11}),
\bfpage{2908}--\blpage{2950}
(\byear{2021})
\end{barticle}
\endbibitem

\bibitem{Shin17}
\begin{bchapter}
\bauthor{\bsnm{Shin}, \binits{H.}},
\bauthor{\bsnm{Lee}, \binits{J.K.}},
\bauthor{\bsnm{Kim}, \binits{J.}},
\bauthor{\bsnm{Kim}, \binits{J.}}:
\bctitle{Continual learning with deep generative replay}.
In: \bbtitle{Advances in Neural Information Processing Systems 30},
pp. \bfpage{2990}--\blpage{2999}
(\byear{2017})
\end{bchapter}
\endbibitem

\bibitem{Wu18_NIPS}
\begin{bchapter}
\bauthor{\bsnm{Wu}, \binits{C.}},
\bauthor{\bsnm{Herranz}, \binits{L.}},
\bauthor{\bsnm{Liu}, \binits{X.}},
\bauthor{\bsnm{wang}, \binits{y.}},
\bauthor{\bparticle{van~de} \bsnm{Weijer}, \binits{J.}},
\bauthor{\bsnm{Raducanu}, \binits{B.}}:
\bctitle{Memory replay gans: Learning to generate new categories without forgetting}.
In: \bbtitle{Advances in Neural Information Processing Systems 31},
pp. \bfpage{5962}--\blpage{5972}
(\byear{2018})
\end{bchapter}
\endbibitem

\bibitem{Bhunia_2022_CVPR}
\begin{bchapter}
\bauthor{\bsnm{Bhunia}, \binits{A.K.}},
\bauthor{\bsnm{Gajjala}, \binits{V.R.}},
\bauthor{\bsnm{Koley}, \binits{S.}},
\bauthor{\bsnm{Kundu}, \binits{R.}},
\bauthor{\bsnm{Sain}, \binits{A.}},
\bauthor{\bsnm{Xiang}, \binits{T.}},
\bauthor{\bsnm{Song}, \binits{Y.-Z.}}:
\bctitle{Doodle it yourself: Class incremental learning by drawing a few sketches}.
In: \bbtitle{Proceedings of the IEEE/CVF Conference on Computer Vision and Pattern Recognition (CVPR)},
pp. \bfpage{2293}--\blpage{2302}
(\byear{2022})
\end{bchapter}
\endbibitem

\bibitem{Hersche_2022_CVPR}
\begin{bchapter}
\bauthor{\bsnm{Hersche}, \binits{M.}},
\bauthor{\bsnm{Karunaratne}, \binits{G.}},
\bauthor{\bsnm{Cherubini}, \binits{G.}},
\bauthor{\bsnm{Benini}, \binits{L.}},
\bauthor{\bsnm{Sebastian}, \binits{A.}},
\bauthor{\bsnm{Rahimi}, \binits{A.}}:
\bctitle{Constrained few-shot class-incremental learning}.
In: \bbtitle{Proceedings of the IEEE/CVF Conference on Computer Vision and Pattern Recognition (CVPR)},
pp. \bfpage{9057}--\blpage{9067}
(\byear{2022})
\end{bchapter}
\endbibitem

\bibitem{Lechun98}
\begin{botherref}
\oauthor{\bsnm{LeChun}, \binits{Y.}}:
The mnist database of handwritten digits
(1998).
\url{http://yann. lecun. com/exdb/mnist/}
\end{botherref}
\endbibitem

\bibitem{bobu21}
\begin{bchapter}
\bauthor{\bsnm{Bobu}, \binits{A.}},
\bauthor{\bsnm{Wiggert}, \binits{M.}},
\bauthor{\bsnm{Tomlin}, \binits{C.}},
\bauthor{\bsnm{Dragan}, \binits{A.D.}}:
\bctitle{Feature expansive reward learning: Rethinking human input}.
In: \bbtitle{Proceedings of the 2021 ACM/IEEE International Conference on Human-Robot Interaction}.
\bsertitle{HRI '21},
pp. \bfpage{216}--\blpage{224}.
\bpublisher{Association for Computing Machinery},
\blocation{New York, NY, USA}
(\byear{2021}).
\doiurl{10.1145/3434073.3444667}.
\burl{https://doi.org/10.1145/3434073.3444667}
\end{bchapter}
\endbibitem

\bibitem{chao10}
\begin{bchapter}
\bauthor{\bsnm{Chao}, \binits{C.}},
\bauthor{\bsnm{Cakmak}, \binits{M.}},
\bauthor{\bsnm{Thomaz}, \binits{A.L.}}:
\bctitle{Transparent active learning for robots}.
In: \bbtitle{2010 5th ACM/IEEE International Conference on Human-Robot Interaction (HRI)},
pp. \bfpage{317}--\blpage{324}
(\byear{2010})
\end{bchapter}
\endbibitem

\bibitem{leite2009time}
\begin{bchapter}
\bauthor{\bsnm{Leite}, \binits{I.}},
\bauthor{\bsnm{Martinho}, \binits{C.}},
\bauthor{\bsnm{Pereira}, \binits{A.}},
\bauthor{\bsnm{Paiva}, \binits{A.}}:
\bctitle{As time goes by: Long-term evaluation of social presence in robotic companions}.
In: \bbtitle{RO-MAN 2009-the 18th IEEE International Symposium on Robot and Human Interactive Communication},
pp. \bfpage{669}--\blpage{674}
(\byear{2009}).
\bcomment{IEEE}
\end{bchapter}
\endbibitem

\bibitem{babel2022will}
\begin{bchapter}
\bauthor{\bsnm{Babel}, \binits{F.}},
\bauthor{\bsnm{Hock}, \binits{P.}},
\bauthor{\bsnm{Kraus}, \binits{J.}},
\bauthor{\bsnm{Baumann}, \binits{M.}}:
\bctitle{It will not take long! longitudinal effects of robot conflict resolution strategies on compliance, acceptance and trust}.
In: \bbtitle{2022 17th ACM/IEEE International Conference on Human-Robot Interaction (HRI)},
pp. \bfpage{225}--\blpage{235}
(\byear{2022}).
\bcomment{IEEE}
\end{bchapter}
\endbibitem

\bibitem{sung2009robots}
\begin{bchapter}
\bauthor{\bsnm{Sung}, \binits{J.}},
\bauthor{\bsnm{Christensen}, \binits{H.I.}},
\bauthor{\bsnm{Grinter}, \binits{R.E.}}:
\bctitle{Robots in the wild: understanding long-term use}.
In: \bbtitle{Proceedings of the 4th ACM/IEEE International Conference on Human Robot Interaction},
pp. \bfpage{45}--\blpage{52}
(\byear{2009})
\end{bchapter}
\endbibitem

\bibitem{Hoffman14}
\begin{barticle}
\bauthor{\bsnm{Hoffman}, \binits{G.}},
\bauthor{\bsnm{Ju}, \binits{W.}}:
\batitle{Designing robots with movement in mind}.
\bjtitle{J. Hum.-Robot Interact.}
\bvolume{3}(\bissue{1}),
\bfpage{91}--\blpage{122}
(\byear{2014}).
\doiurl{10.5898/JHRI.3.1.Hoffman}
\end{barticle}
\endbibitem

\bibitem{anzalone__2015}
\begin{barticle}
\bauthor{\bsnm{Anzalone}, \binits{S.M.}},
\bauthor{\bsnm{Boucenna}, \binits{S.}},
\bauthor{\bsnm{Ivaldi}, \binits{S.}},
\bauthor{\bsnm{Chetouani}, \binits{M.}}:
\batitle{Evaluating the engagement with social robots}.
\bjtitle{International Journal of Social Robotics}
\bvolume{7}(\bissue{4}),
\bfpage{465}--\blpage{478}
(\byear{2015}).
\doiurl{10.1007/s12369-015-0298-7}
\end{barticle}
\endbibitem

\bibitem{takayama11}
\begin{bchapter}
\bauthor{\bsnm{Takayama}, \binits{L.}},
\bauthor{\bsnm{Dooley}, \binits{D.}},
\bauthor{\bsnm{Ju}, \binits{W.}}:
\bctitle{Expressing thought: Improving robot readability with animation principles}.
In: \bbtitle{2011 6th ACM/IEEE International Conference on Human-Robot Interaction (HRI)},
pp. \bfpage{69}--\blpage{76}
(\byear{2011})
\end{bchapter}
\endbibitem

\bibitem{aliasghari21}
\begin{bchapter}
\bauthor{\bsnm{Aliasghari}, \binits{P.}},
\bauthor{\bsnm{Ghafurian}, \binits{M.}},
\bauthor{\bsnm{Nehaniv}, \binits{C.L.}},
\bauthor{\bsnm{Dautenhahn}, \binits{K.}}:
\bctitle{Effects of gaze and arm motion kinesics on a humanoid's perceived confidence, eagerness to learn, and attention to the task in a teaching scenario}.
In: \bbtitle{Proceedings of the 2021 ACM/IEEE International Conference on Human-Robot Interaction}.
\bsertitle{HRI '21},
pp. \bfpage{197}--\blpage{206}.
\bpublisher{Association for Computing Machinery},
\blocation{New York, NY, USA}
(\byear{2021}).
\doiurl{10.1145/3434073.3444651}.
\burl{https://doi.org/10.1145/3434073.3444651}
\end{bchapter}
\endbibitem

\bibitem{robinette17}
\begin{barticle}
\bauthor{\bsnm{Robinette}, \binits{P.}},
\bauthor{\bsnm{Howard}, \binits{A.M.}},
\bauthor{\bsnm{Wagner}, \binits{A.R.}}:
\batitle{Effect of robot performance on human–robot trust in time-critical situations}.
\bjtitle{IEEE Transactions on Human-Machine Systems}
\bvolume{47}(\bissue{4}),
\bfpage{425}--\blpage{436}
(\byear{2017}).
\doiurl{10.1109/THMS.2017.2648849}
\end{barticle}
\endbibitem

\bibitem{rossi20}
\begin{bchapter}
\bauthor{\bsnm{Rossi}, \binits{A.}},
\bauthor{\bsnm{Dautenhahn}, \binits{K.}},
\bauthor{\bsnm{Koay}, \binits{K.L.}},
\bauthor{\bsnm{Walters}, \binits{M.L.}},
\bauthor{\bsnm{Holthaus}, \binits{P.}}:
\bctitle{Evaluating people's perceptions of trust in a robot in a repeated interactions study}.
In: \beditor{\bsnm{Wagner}, \binits{A.R.}},
\beditor{\bsnm{Feil-Seifer}, \binits{D.}},
\beditor{\bsnm{Haring}, \binits{K.S.}},
\beditor{\bsnm{Rossi}, \binits{S.}},
\beditor{\bsnm{Williams}, \binits{T.}},
\beditor{\bsnm{He}, \binits{H.}},
\beditor{\bsnm{Sam~Ge}, \binits{S.}} (eds.)
\bbtitle{Social Robotics},
pp. \bfpage{453}--\blpage{465}.
\bpublisher{Springer},
\blocation{Cham}
(\byear{2020})
\end{bchapter}
\endbibitem

\bibitem{gadre19}
\begin{bchapter}
\bauthor{\bsnm{Gadre}, \binits{S.Y.}},
\bauthor{\bsnm{Rosen}, \binits{E.}},
\bauthor{\bsnm{Chien}, \binits{G.}},
\bauthor{\bsnm{Phillips}, \binits{E.}},
\bauthor{\bsnm{Tellex}, \binits{S.}},
\bauthor{\bsnm{Konidaris}, \binits{G.}}:
\bctitle{End-user robot programming using mixed reality}.
In: \bbtitle{2019 International Conference on Robotics and Automation (ICRA)},
pp. \bfpage{2707}--\blpage{2713}
(\byear{2019}).
\doiurl{10.1109/ICRA.2019.8793988}
\end{bchapter}
\endbibitem

\bibitem{solanes20}
\begin{barticle}
\bauthor{\bsnm{Solanes}, \binits{J.E.}},
\bauthor{\bsnm{Muñoz}, \binits{A.}},
\bauthor{\bsnm{Gracia}, \binits{L.}},
\bauthor{\bsnm{Martí}, \binits{A.}},
\bauthor{\bsnm{Girbés-Juan}, \binits{V.}},
\bauthor{\bsnm{Tornero}, \binits{J.}}:
\batitle{Teleoperation of industrial robot manipulators based on augmented reality}.
\bjtitle{The International Journal of Advanced Manufacturing Technology}
\bvolume{111}(\bissue{3-4}),
\bfpage{1077}--\blpage{1097}
(\byear{2020}).
\doiurl{10.1007/s00170-020-05997-1}
\end{barticle}
\endbibitem

\bibitem{louie20}
\begin{barticle}
\bauthor{\bsnm{Louie}, \binits{W.-Y.G.}},
\bauthor{\bsnm{Nejat}, \binits{G.}}:
\batitle{A social robot learning to facilitate an assistive group-based activity from non-expert caregivers}.
\bjtitle{International Journal of Social Robotics}
\bvolume{12}(\bissue{5}),
\bfpage{1159}--\blpage{1176}
(\byear{2020}).
\doiurl{10.1007/s12369-020-00621-4}
\end{barticle}
\endbibitem

\bibitem{schrumtowards}
\begin{botherref}
\oauthor{\bsnm{Schrum}, \binits{M.L.}},
\oauthor{\bsnm{Hedlund-Botti}, \binits{E.}},
\oauthor{\bsnm{Gombolay}, \binits{M.C.}}:
Towards Improving Life-Long Learning Via Personalized, Reciprocal Teaching.
Workshop on Lifelong Learning and Personalization in Long-Term Human-Robot Interaction (LEAP-HRI), 17th ACM/IEEE International Conference on Human-Robot Interaction (HRI)
(2022)
\end{botherref}
\endbibitem

\bibitem{Liu_2020}
\begin{barticle}
\bauthor{\bsnm{Liu}, \binits{B.}}:
\batitle{Learning on the job: Online lifelong and continual learning}.
\bjtitle{Proceedings of the AAAI Conference on Artificial Intelligence}
\bvolume{34}(\bissue{09}),
\bfpage{13544}--\blpage{13549}
(\byear{2020}).
\doiurl{10.1609/aaai.v34i09.7079}
\end{barticle}
\endbibitem

\bibitem{Russakovsky15}
\begin{barticle}
\bauthor{\bsnm{Russakovsky}, \binits{O.}},
\bauthor{\bsnm{Deng}, \binits{J.}},
\bauthor{\bsnm{Su}, \binits{H.}},
\bauthor{\bsnm{Krause}, \binits{J.}},
\bauthor{\bsnm{Satheesh}, \binits{S.}},
\bauthor{\bsnm{Ma}, \binits{S.}},
\bauthor{\bsnm{Huang}, \binits{Z.}},
\bauthor{\bsnm{Karpathy}, \binits{A.}},
\bauthor{\bsnm{Khosla}, \binits{A.}},
\bauthor{\bsnm{Bernstein}, \binits{M.}},
\bauthor{\bsnm{Berg}, \binits{A.C.}},
\bauthor{\bsnm{Fei-Fei}, \binits{L.}}:
\batitle{Imagenet large scale visual recognition challenge}.
\bjtitle{Int. J. Comput. Vision}
\bvolume{115}(\bissue{3}),
\bfpage{211}--\blpage{252}
(\byear{2015})
\end{barticle}
\endbibitem

\bibitem{Redmon_2016_CVPR}
\begin{bchapter}
\bauthor{\bsnm{Redmon}, \binits{J.}},
\bauthor{\bsnm{Divvala}, \binits{S.}},
\bauthor{\bsnm{Girshick}, \binits{R.}},
\bauthor{\bsnm{Farhadi}, \binits{A.}}:
\bctitle{You only look once: Unified, real-time object detection}.
In: \bbtitle{Proceedings of the IEEE Conference on Computer Vision and Pattern Recognition (CVPR)}
(\byear{2016})
\end{bchapter}
\endbibitem

\bibitem{Hosang_2017_CVPR}
\begin{bchapter}
\bauthor{\bsnm{Hosang}, \binits{J.}},
\bauthor{\bsnm{Benenson}, \binits{R.}},
\bauthor{\bsnm{Schiele}, \binits{B.}}:
\bctitle{Learning non-maximum suppression}.
In: \bbtitle{Proceedings of the IEEE Conference on Computer Vision and Pattern Recognition (CVPR)}
(\byear{2017})
\end{bchapter}
\endbibitem

\bibitem{chai18_ijcai}
\begin{bchapter}
\bauthor{\bsnm{Chai}, \binits{J.Y.}},
\bauthor{\bsnm{Gao}, \binits{Q.}},
\bauthor{\bsnm{She}, \binits{L.}},
\bauthor{\bsnm{Yang}, \binits{S.}},
\bauthor{\bsnm{Saba-Sadiya}, \binits{S.}},
\bauthor{\bsnm{Xu}, \binits{G.}}:
\bctitle{Language to action: Towards interactive task learning with physical agents}.
In: \bbtitle{Proceedings of the Twenty-Seventh International Joint Conference on Artificial Intelligence, {IJCAI-18}},
pp. \bfpage{2}--\blpage{9}.
\bpublisher{International Joint Conferences on Artificial Intelligence Organization},
\blocation{Stockholm}
(\byear{2018}).
\doiurl{10.24963/ijcai.2018/1}
\end{bchapter}
\endbibitem

\bibitem{qualtrics}
\begin{botherref}
Qualtrics.
\url{https://www.qualtrics.com}
(2005)
\end{botherref}
\endbibitem

\bibitem{Madsen00HCT}
\begin{bchapter}
\bauthor{\bsnm{Madsen}, \binits{M.}},
\bauthor{\bsnm{Gregor}, \binits{S.}}:
\bctitle{Measuring human-computer trust}.
In: \bbtitle{Proceedings of the 11 Th Australasian Conference on Information Systems},
pp. \bfpage{6}--\blpage{8}
(\byear{2000})
\end{bchapter}
\endbibitem

\bibitem{Carpinella17RoSAS}
\begin{bchapter}
\bauthor{\bsnm{Carpinella}, \binits{C.M.}},
\bauthor{\bsnm{Wyman}, \binits{A.B.}},
\bauthor{\bsnm{Perez}, \binits{M.A.}},
\bauthor{\bsnm{Stroessner}, \binits{S.J.}}:
\bctitle{The {Robotic} {Social} {Attributes} {Scale} ({RoSAS}): {Development} and {Validation}}.
In: \bbtitle{Proceedings of the 2017 {ACM}/{IEEE} {International} {Conference} on {Human}-{Robot} {Interaction}},
pp. \bfpage{254}--\blpage{262}.
\bpublisher{ACM},
\blocation{Vienna Austria}
(\byear{2017}).
\doiurl{10.1145/2909824.3020208}.
\burl{https://dl.acm.org/doi/10.1145/2909824.3020208}
Accessed 2022-09-29
\end{bchapter}
\endbibitem

\bibitem{Hart06TLX}
\begin{barticle}
\bauthor{\bsnm{Hart}, \binits{S.G.}}:
\batitle{Nasa-{Task} {Load} {Index} ({NASA}-{TLX}); 20 {Years} {Later}}.
\bjtitle{Proceedings of the Human Factors and Ergonomics Society Annual Meeting}
\bvolume{50}(\bissue{9}),
\bfpage{904}--\blpage{908}
(\byear{2006}).
\doiurl{10.1177/154193120605000909}.
Accessed 2022-09-29
\end{barticle}
\endbibitem

\bibitem{Brooke95SUS}
\begin{botherref}
\oauthor{\bsnm{Brooke}, \binits{J.}}:
Sus: A quick and dirty usability scale.
Usability Eval. Ind.
\textbf{189}
(1995)
\end{botherref}
\endbibitem

\bibitem{shapiro1965analysis}
\begin{barticle}
\bauthor{\bsnm{Shapiro}, \binits{S.S.}},
\bauthor{\bsnm{Wilk}, \binits{M.B.}}:
\batitle{An analysis of variance test for normality (complete samples)}.
\bjtitle{Biometrika}
\bvolume{52}(\bissue{3/4}),
\bfpage{591}--\blpage{611}
(\byear{1965})
\end{barticle}
\endbibitem

\bibitem{Wilcoxon45Test}
\begin{barticle}
\bauthor{\bsnm{Wilcoxon}, \binits{F.}}:
\batitle{Individual comparisons by ranking methods}.
\bjtitle{Biometrics Bulletin}
\bvolume{1}(\bissue{6}),
\bfpage{80}--\blpage{83}
(\byear{1945})
\end{barticle}
\endbibitem

\bibitem{armstrong2014use}
\begin{barticle}
\bauthor{\bsnm{Armstrong}, \binits{R.A.}}:
\batitle{When to use the {B}onferroni correction}.
\bjtitle{Ophthalmic and Physiological Optics}
\bvolume{34}(\bissue{5}),
\bfpage{502}--\blpage{508}
(\byear{2014})
\end{barticle}
\endbibitem

\end{thebibliography}


\begin{appendices}
\section{FT as a Baseline Approach for the Study}
\label{sec:ft_justify}
It can be argued that FT is a sub-optimal model that users would perceive negatively, and thus it should not be tested. However, although it is expected that users might perceive FT negatively, it is unknown if there is a statistically significant improvement in human perceptions when the robot uses a SOTA CL model (CBCL). Therefore, to quantify human perceptions of CL models, it is necessary to compare them against a baseline (FT) in the context of CL through HRI.
Further, a unique aspect of FT is that, although it forgets objects from previous sessions, it can accurately classify new objects. In the context of our study, it is unknown if users would even care about the performance of the model on old objects and perhaps be more forgiving of the model as it is able to learn new objects.
Note that in SGCL participants are not told to test the robot on old objects. Instead, they have flexibility regarding which objects to test in each session. Results for FT (Section \ref{sec:results}) show the effect of this flexibility. 

Finally, if we chose a baseline that simply does not learn anything, that would be a trivial comparison, since users would be expected to perceive this model negatively. Another choice of a baseline model is to pre-train a model on a set of objects. However, our study is not only about CL but also about human-robot teaching in an unconstrained manner. In our study, users are given high flexibility in teaching the objects. Particularly, users can name the objects however they like, teach an object as many times as they like in a session, and bring in a few of their own objects. In such an unconstrained environment, it would be impossible to choose a baseline that is pre-trained on some objects. For these reasons, we believe it is important to test the FT model in our study as a baseline.

\begin{figure}[t]
\centering
\includegraphics[width=0.45\linewidth]{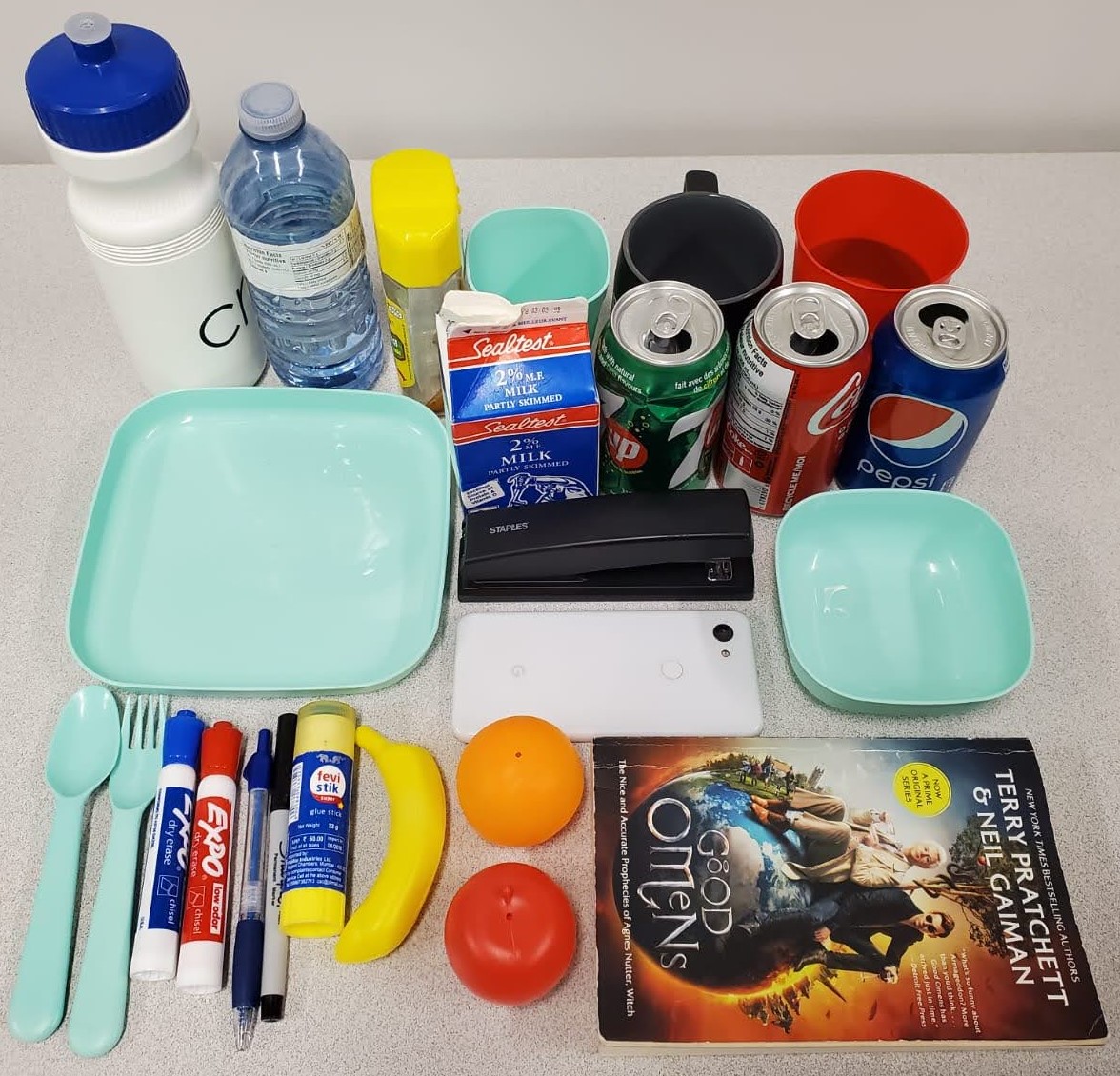}
\caption{\small The twenty-five objects used in our study. Note that the visual similarity of some objects and their size variation make this a challenging task.}
\label{fig:objects}
\end{figure}

\section{Objects used in the Study}\label{secA1}
Figure \ref{fig:objects} shows the twenty-five objects used in our study. We used realistic objects from daily life with significant variations in size, color, and shape, which made the object-learning task quite challenging. Participants taught these objects in five sessions with five objects per session. Participants were also allowed to bring objects of their own choice after the first two sessions. If the participants brought their own objects, we used them to  replace some of the twenty-five objects. Overall, each participant taught twenty-five objects. 

\section{Overall Results of the Study}
Table \ref{tab:overall_results} shows the overall statistics of the study across all 60 participants. Tables \ref{tab:trust_detailed}, \ref{tab:rosas_detailed}, \ref{tab:task_load_detailed}, \ref{tab:usability_detailed} show the overall results for the four questionnaires for the three conditions across all 60 participants. 

\begin{table*}
\centering
\caption{Overall results of the study irrespective of the CL model chosen.}
{%
\newcommand{\mc}[3]{\multicolumn{#1}{#2}{#3}}
\begin{center}
\begin{tabular}{P{1.0cm}P{0.5cm}P{0.5cm}P{0.5cm}P{0.5cm}P{.5cm}P{.5cm}P{.5cm}P{.5cm}P{.5cm}P{.5cm}}
Session & \mc{2}{c}{1} & \mc{2}{c}{2} & \mc{2}{c}{3} & \mc{2}{c}{4} & \mc{2}{c}{5} 
\\
Value & $\mu$ & $\sigma$ & $\mu$ & $\sigma$ & $\mu$ & $\sigma$ & $\mu$ & $\sigma$ & $\mu$ & $\sigma$ 
\\ 
\cline{1-11}
Trust & 2.81 & 0.86 & 2.68 & 0.88 & 2.31 & 0.89 & 2.29 & 0.88 & 2.29 & 0.92\\ 
Warm. & 2.93 & 1.34 & × & × & × & × & × & × & 2.71 & 1.36 \\
Comp. & 4.91 & 1.30 & × & × & × & × & × & × & 4.14 & 1.39 \\
Disc. & 1.88 & 0.89 & × & × & × & × & × & × & 1.98 & 0.91 \\
TLX & 26.5 & 9.63 & 25.7 & 11.4 & 26.8 & 12.2 & 27.6 & 13.4 & 27.5 & 14.3 \\
SUS & 74.4 & 16.8 & 72.9 & 17.1 & 68.9 & 15.5 & 67.9 & 18.8 & 67.5 & 16.3 \\
\end{tabular}

\end{center}
}%
\label{tab:overall_results}
\end{table*}

\begin{table*}
\centering
\caption{Detailed results for trust in the three conditions. NS stands for not significant.}
{%
\newcommand{\mc}[3]{\multicolumn{#1}{#2}{#3}}
\begin{center}

\begin{tabular}{P{1.0cm}P{0.7cm}P{0.7cm}P{0.7cm}P{0.7cm}P{0.7cm}P{0.7cm}}
\mc{1}{c}{Session} & \mc{2}{c}{CBCL} & \mc{2}{c}{FT} & \mc{2}{c}{JT}\\
Value & $\mu$ & $\sigma$ & $\mu$ & $\sigma$ & $\mu$ & $\sigma$\\ 
\cline{1-7}
1 & 3.07 & 0.53 & 2.52 & 0.96 & 2.85 & 0.95\\
2 & 3.09 & 0.49 & 1.92 & 0.79 & 2.99 & 0.80\\
3 & 2.76 & 0.67 & 1.66 & 0.81 & 2.49 & 0.82\\
4 & 2.64 & 0.68 & 1.75 & 0.88 & 2.41 & 0.87\\
5 & 2.74 & 0.77 & 1.51 & 0.66 & 2.61 & 0.82\\
all & 2.86 & 0.65 & 1.88 & 0.89 & 2.67 & 0.87\\
\cline{1-7}
\end{tabular}
\\[0.7em]

\begin{tabular}{P{1.0cm}P{1.8cm}P{1.0cm} P{0.7cm}P{0.7cm} P{1.7cm}P{0.9cm}}
\mc{1}{c}{Session} & \mc{2}{c}{FT-CBCL} & \mc{2}{c}{CBCL-JT} & \mc{2}{c}{FT-JT}\\
Value & $p$ & $W$ & $p$ & $W$ & $p$ & $W$\\ 
\cline{1-7}
1 & 0.0230 & 130 & NS & NS & NS & NS\\
2 & 3.3$\times 10^{-5}$ & 42.5 & NS & NS & 0.0003 & 316.5\\
3 & 0.0001 & 62.5 & NS & NS & 0.0046 & 305 \\
4 & 0.0030 & 83.5 & NS & NS & 0.0369 & 240 \\
5 & 6.3$\times 10^{-5}$ & 56.5 & NS & NS & 0.0001 & 354\\
all & 2.2$\times 10^{-14}$ & 1930.5 & NS & NS & 3.5$\times 10^{-9}$ & 7344\\
\cline{1-7}
\end{tabular}
\end{center}
}%
\label{tab:trust_detailed}
\end{table*}

\begin{table*}
\centering
\caption{Detailed results for usability in the three conditions. NS stands for not significant.}
{%
\newcommand{\mc}[3]{\multicolumn{#1}{#2}{#3}}
\begin{center}
\begin{tabular}{P{1.0cm}P{0.7cm}P{0.7cm}P{0.7cm}P{0.7cm}P{0.7cm}P{0.7cm}}
\mc{1}{c}{Session} & \mc{2}{c}{CBCL} & \mc{2}{c}{FT} & \mc{2}{c}{JT}\\
Value & $\mu$ & $\sigma$ & $\mu$ & $\sigma$ & $\mu$ & $\sigma$\\ 
\cline{1-7}
1 & 75.9 & 20.6 & 73.3 & 14.0 & 73.9 & 15.6\\
2 & 77.4 & 11.3 & 64.5 & 21.2 & 76.6 & 14.9\\
3 & 71.9 & 17.5 & 62.6 & 15.9 & 72 & 10.9\\
4 & 73.5 & 16.5 & 58.2 & 22.3 & 70.9 & 14.5\\
5 & 72.7 & 16.7 & 59.0 & 17.4 & 70.2 & 11.7\\
all & 74.3 & 16.6 & 63.7 & 18.8 & 72.7 & 13.6\\
\cline{1-7}
\end{tabular}
\\[0.7em]

\begin{tabular}{P{1.0cm}P{1.8cm}P{1.0cm} P{0.7cm}P{0.7cm} P{1.7cm}P{0.9cm}}
\mc{1}{c}{Session} & \mc{2}{c}{FT-CBCL} & \mc{2}{c}{CBCL-JT} & \mc{2}{c}{FT-JT}\\
Value & $p$ & $W$ & $p$ & $W$ & $p$ & $W$\\ 
\cline{1-7}
1 & NS & NS & NS & NS & NS & NS\\
2 & 0.0386 & 130.5 & NS & NS & NS & NS\\
3 & NS & NS & NS & NS & 0.0321 & 279.5\\
4 & 0.0353 & 114 & NS & NS & 0.0442 & 237.5\\
5 & 0.0145 & 116 & NS & NS & 0.02624 & 295.5\\
all & 2.0$\times 10^{-5}$ & 3441 & NS & NS & 0.0002 & 6521\\
\cline{1-7}
\end{tabular}
\end{center}
}%
\label{tab:usability_detailed}
\end{table*}

\begin{table*}
\centering
\caption{Detailed results for task load in the three conditions. No statistically significant difference was seen for any condition, therefore $p$ and $W$ values are not reported.}
{%
\newcommand{\mc}[3]{\multicolumn{#1}{#2}{#3}}
\begin{center}
\begin{tabular}{P{1.0cm}P{0.7cm}P{0.7cm}P{0.7cm}P{0.7cm}P{0.7cm}P{0.7cm}}
\mc{1}{c}{Session} & \mc{2}{c}{CBCL} & \mc{2}{c}{FT} & \mc{2}{c}{JT}\\
Value & $\mu$ & $\sigma$ & $\mu$ & $\sigma$ & $\mu$ & $\sigma$\\ 
\cline{1-7}
1 & 24.4 & 7.94 & 28.9 & 9.72 & 27.2 & 11.7\\
2 & 25.1 & 13.3 & 29.5 & 11.5 & 22.9 & 7.98\\
3 & 25.8 & 11.9 & 27.9 & 14.6 & 27.1 & 10.9\\
4 & 28.1 & 14.3 & 26.8 & 14.7 & 27.5 & 11.9\\
5 & 28.3 & 16.8 & 27.9 & 16.2 & 26.4 & 9.38\\
all & 26.4 & 13.0 & 28.3 & 13.3 & 26.2 & 10.3\\
\cline{1-7}
\end{tabular}
\end{center}
}%
\label{tab:task_load_detailed}
\end{table*}

\begin{table*}
\centering
\caption{Detailed results for robot's social attributes (warmth, competence, discomfort) in the three conditions. NS stands for not significant.}
{%
\newcommand{\mc}[3]{\multicolumn{#1}{#2}{#3}}
\begin{center}
\begin{tabular}{P{1.9cm}P{1.0cm}P{0.7cm}P{0.7cm}P{0.7cm}P{0.7cm}P{0.7cm}P{0.7cm}}
 & \mc{1}{c}{Session} & \mc{2}{c}{CBCL} & \mc{2}{c}{FT} & \mc{2}{c}{JT}\\
 & Value & $\mu$ & $\sigma$ & $\mu$ & $\sigma$ & $p$ & $W$\\ 
\cline{1-8}
\multirow{3}{*}{Warmth} & 1 & 2.62 & 1.19 & 3.22 & 1.35 & 2.97 & 1.46\\
 & 5 & 2.69 & 1.46 & 2.55 & 1.28 & 2.86 & 1.37\\
 & all & 2.65 & 1.32 & 2.89 & 1.35 & 2.92 & 1.40\\ 
\cline{1-8}
\multirow{3}{*}{Competence} & 1 & 5.04 & 1.12 & 4.72 & 1.39 & 4.97 & 1.41\\
 & 5 & 4.49 & 1.41 & 3.27 & 1.17 & 4.63 & 1.19\\
 & all & 4.77 & 1.29 & 4.01 & 1.47 & 4.8 & 1.30\\
\cline{1-8}
\multirow{3}{*}{Discomfort} & 1 & 1.91 & 0.89 & 1.99 & 0.83 & 1.74 & 0.97\\
 & 5 & 2.02 & 0.82 & 2.00 & 0.92 & 1.94 & 1.01\\
 & all & 1.96 & 0.84 & 1.99 & 0.86 & 1.84 & 0.98\\
 \cline{1-8}
\end{tabular}
\\[0.7em]

\begin{tabular}{P{1.3cm}P{1.0cm}P{1.5cm}P{0.7cm}P{0.7cm}P{0.7cm}P{1.5cm}P{0.7cm}}
 & \mc{1}{c}{Session} & \mc{2}{c}{FT-CBCL} & \mc{2}{c}{CBCL-JT} & \mc{2}{c}{FT-JT}\\
 & Value & $p$ & $W$ & $p$ & $W$ & $p$ & $W$\\ 
\cline{1-8}
\multirow{3}{*}{Warmth} & 1 & NS & NS & NS & NS & NS & NS\\
 & 5 & NS & NS & NS & NS & NS & NS\\
 & all & NS & NS & NS & NS & NS & NS\\ 
\cline{1-8}
\multirow{3}{*}{Competence} & 1 & NS & NS & NS & NS & NS & NS\\
 & 5 & 0.0065 & 105.5 & NS & NS & 0.0013 & 333.5\\
 & all & 0.0132 & 604.5 & NS & NS & 0.0155 & 1127\\
\cline{1-8}
\multirow{3}{*}{Discomfort} & 1 & NS & NS & NS & NS & NS & NS\\
 & 5 & NS & NS & NS & NS & NS & NS\\
 & all & NS & NS & NS & NS & NS & NS\\
 \cline{1-8}
\end{tabular}

\end{center}
}%
\label{tab:rosas_detailed}
\end{table*}

\end{appendices}

\end{document}